\documentclass{article} 
\usepackage{collas2026_conference,times}
\usepackage{easyReview}

\usepackage{amsmath,amsfonts,bm}

\def\eqref#1{equation~\ref{#1}}

\def\1{\bm{1}}

\DeclareMathAlphabet{\mathsfit}{\encodingdefault}{\sfdefault}{m}{sl}
\SetMathAlphabet{\mathsfit}{bold}{\encodingdefault}{\sfdefault}{bx}{n}

\usepackage{hyperref}
\hypersetup{
    colorlinks=true,
    linkcolor=mydarkblue,
    filecolor=magenta,
    urlcolor=blue,
    citecolor=purple,
    pdftitle={Overleaf Example},
    pdfpagemode=FullScreen,
    }

\usepackage{graphicx}
\usepackage{subcaption}
\usepackage{mathtools}
\usepackage{microtype}
\usepackage{float}
\usepackage{url}
\usepackage{booktabs}
\usepackage{enumitem}
\usepackage{verbatim}
\usepackage{amsthm}
\usepackage{amsmath}
\usepackage{amssymb}
\usepackage{tikz}
\usepackage[most]{tcolorbox}
\usepackage[dvipsnames]{xcolor}
\usepackage{xcolor}
\usepackage{tabularx}
\usepackage{makecell}
\usepackage{titletoc}
\usepackage{etoolbox}
\usepackage{aliascnt}
\usepackage{algorithm}
\usepackage{algpseudocode}
\usepackage{silence}

\newcolumntype{Y}{>{\raggedright\arraybackslash}X}

\newcommand{\TriComment}[1]{\hfill $\triangleright$ #1}

\newcommand{\indep}{\mathrel{\perp\mspace{-10mu}\perp}}

\setlist{nolistsep}

\newcommand\blfootnote[1]{%
  \begingroup
  \renewcommand\thefootnote{}\footnote{#1}%
  \addtocounter{footnote}{-1}%
  \endgroup
}

\newcommand{\spaceddotfill}{%
  \hspace{0.25em}
  \leavevmode
  \cleaders\hbox to 1em{\hss.\hss}\hfill
  \kern0pt%
}

\definecolor{mydarkblue}{rgb}{0,0.08,0.45}

\usepackage{tikz}
\usetikzlibrary{positioning, arrows.meta}

\usepackage[capitalize,noabbrev, nameinlink]{cleveref}
\theoremstyle{plain}
\newtheorem{theorem}{Theorem}[section]

\newtheorem{assumption}{Assumption}[section]
\newtheorem{lemma}{Lemma}[section]
\newaliascnt{corollary}{theorem}

\aliascntresetthe{corollary}
\newaliascnt{proposition}{theorem}

\aliascntresetthe{proposition}
\theoremstyle{definition}
\newaliascnt{definition}{theorem}
\newtheorem{definition}[definition]{Definition}
\aliascntresetthe{definition}
\theoremstyle{remark}
\newaliascnt{remark}{theorem}

\aliascntresetthe{remark}
\newaliascnt{example}{theorem}

\aliascntresetthe{example}
\newaliascnt{desideratum}{theorem}
\newtheorem{desideratum}[desideratum]{Desideratum}
\aliascntresetthe{desideratum}
\newaliascnt{notation}{theorem}

\aliascntresetthe{notation}
\newtheoremstyle{scenario}
  {3pt}
  {3pt}
  {\normalfont}
  {}
  {\bfseries}
  {.}
  {.5em}
  {\thmname{#1}\thmnumber{ #2}\thmnote{ \textup{\textit{(#3)}}}}
\theoremstyle{scenario}
\newaliascnt{scenario}{theorem}
\newtheorem{scenario}[scenario]{Scenario}
\aliascntresetthe{scenario}

\newcommand{\theoremanchor}[1]{theorem:#1}
\newcommand{\theoremnumnumber}[1]{theoremnum:#1}
\newenvironment{theoremnum}[3][]
  {\par\medskip
   \expandafter\gdef\csname\theoremnumnumber{#2}\endcsname{#3}%
   \phantomsection
   \hypertarget{\theoremanchor{#2}}{}%
   \noindent
   \textbf{Theorem #3}%
   \if\relax\detokenize{#1}\relax
     \textbf{.}%
   \else
     \textup{ (#1).}%
   \fi
   \itshape}
  {\par\medskip}

\newcommand{\theoremnumref}[1]{%
  \hyperlink{\theoremanchor{#1}}%
  {Theorem~\csname\theoremnumnumber{#1}\endcsname}}

\tcbuselibrary{theorems}

\tcbset{
  thmstyle/.style={
    enhanced,
    fonttitle=\bfseries,
    coltitle=black,
    boxrule=0.8pt,
    arc=2mm,
    left=1mm, right=1mm, top=1mm, bottom=1mm,
  }
}

\newtheorem{takeaway}{Takeaway}
\newtheorem*{openquestion}{Open Question}
\tcolorboxenvironment{summary}{thmstyle, colback=orange!5!white, colframe=orange!80!black}
\tcolorboxenvironment{takeaway}{thmstyle, colback=blue!5!white, colframe=blue!75!black}
\tcolorboxenvironment{openquestion}{thmstyle, colback=violet!5!white, colframe=violet!75!black}

\makeatletter
\crefname{section}{\S\@gobble}{\S\S}
\crefname{subsection}{\S\@gobble}{\S\S}
\makeatother
\crefname{appendix}{\S}{\S\S}
\crefname{figure}{Figure}{Figures}
\crefname{table}{Table}{Tables}
\crefname{theorem}{Theorem}{Theorems}
\crefname{informal_theorem}{Theorem (Informal)}{Theorems (Informal)}
\crefname{assumption}{Assumption}{Assumptions}
\crefname{lemma}{Lemma}{Lemmas}
\crefname{corollary}{Corollary}{Corollaries}
\crefname{proposition}{Proposition}{Propositions}
\crefname{definition}{Definition}{Definitions}
\crefname{remark}{Remark}{Remarks}
\crefname{example}{Example}{Examples}
\crefname{desideratum}{Desideratum}{Desiderata}
\crefname{notation}{Notation}{Notations}
\crefname{scenario}{Scenario}{Scenarios}
\Crefname{theorem}{Theorem}{Theorems}
\Crefname{informal_theorem}{Theorem (Informal)}{Theorems}
\Crefname{assumption}{Assumption}{Assumptions}
\Crefname{lemma}{Lemma}{Lemmas}
\Crefname{corollary}{Corollary}{Corollaries}
\Crefname{proposition}{Proposition}{Propositions}
\Crefname{definition}{Definition}{Definitions}
\Crefname{remark}{Remark}{Remarks}
\Crefname{example}{Example}{Examples}
\Crefname{desideratum}{Desideratum}{Desiderata}
\Crefname{notation}{Notation}{Notations}
\Crefname{scenario}{Scenario}{Scenarios}
\renewcommand{\eqref}[1]{(\ref{#1})}
\crefrangeformat{assumption}{%
  Assumptions~#3#1#4--#5#2#6%
}

\title{Forgetting is Everywhere}

\author{
  \parbox{\linewidth}{\centering
    Ben Sanati\textsuperscript{1}\qquad Thomas L. Lee\textsuperscript{1}\qquad Trevor McInroe\textsuperscript{1}\qquad Aidan Scannell\textsuperscript{1}\\
    \textbf{Esmeralda S. Whitammer}\textsuperscript{1,2}\quad David Abel\textsuperscript{1}\quad Amos Storkey\textsuperscript{1}
  }
}

\collasfinalcopy 

\begin{document}

\maketitle

\begin{abstract}
  \looseness=-1
    A fundamental challenge in developing general learning algorithms is their tendency to forget past knowledge as they adapt to new data.
    Addressing this problem requires a principled understanding of forgetting.
    Yet, despite decades of study, no unified definition has emerged that offers insight into the underlying dynamics of learning.
    We propose an algorithm- and task-agnostic theory that characterises forgetting as a lack of self-consistency in a learner's predictive distribution, manifesting as a loss of predictive information.
    Our theory naturally yields a general measure of an algorithm's propensity to forget, proves that exact Bayesian inference allows for adaptation without forgetting, and provides a tautological explanation for why generative models forget when trained on their own synthetic outputs.
    To validate these claims, we design a comprehensive set of experiments that span classification, regression, generative modelling, and reinforcement learning.
    We demonstrate that forgetting is present across all deep learning settings and plays a significant role in determining learning efficiency.
    Together, these results establish a principled understanding of forgetting and lay the foundation for analysing and improving the information retention capabilities of general learning algorithms.
\blfootnote{\textsuperscript{1}Institute for Machine Learning, University of Edinburgh. \textsuperscript{2}CIFAR Fellow. Correspondence to: b.sanati@ed.ac.uk.}
\end{abstract}

\section{Introduction}
\looseness=-1
Forgetting is a ubiquitous yet poorly understood phenomenon in machine learning \citep{mccloskey1989catastrophic}. When a learner updates its beliefs based on new observations, information acquired from previous observations is often lost. Although this behaviour is well documented in non-stationary settings such as continual learning (CL; \citeauthor{kirkpatrick2017overcoming}, \citeyear{kirkpatrick2017overcoming}) and reinforcement learning (RL; \citeauthor{atkinson2021pseudo}, \citeyear{atkinson2021pseudo}; \citeauthor{khetarpal2022towards}, \citeyear{khetarpal2022towards}), forgetting is present even in independent and identically distributed (i.i.d.) CL settings without task shift \citep{chunking}.

\looseness=-1
Most studies of forgetting come from the CL literature. Here, metrics typically track how performance degrades on earlier tasks after training on later tasks \citep{chaudhry2018riemannian}.
Such metrics rarely distinguish between constructive adaptation (assimilation of new tasks) and destructive adaptation (loss of capability on past tasks). Therefore, they conflate two distinct phenomena: \emph{backward transfer}, in which new learning improves performance on past tasks \citep{benavides2022theory}, and \emph{forgetting}, in which new updates erode prior knowledge \citep{jagielski2022measuring}.

\looseness=-1
In contrast, we study forgetting as a \emph{fundamental property of learning}; a consequence of how any adaptive system updates its beliefs.
As a consequence of this generality, we can study the role of forgetting in learning abstractly and accommodate specific choices of problem settings or learning algorithms as instances of this general framing.

Our new conceptual foundation of forgetting is built on the following insight:
\begin{quote}
  \vspace*{-0.25em}
    \emph{\looseness=-1
    A learning algorithm admits a predictive representation of the agent's state. Forgetting can be evaluated by whether the learner's update procedure preserves consistency in this predictive representation.
    }
  \vspace*{-1em}
\end{quote}
We ask: \emph{What is forgetting? When and why does it occur? How does it impact learning?} Wielding the above insight, we provide a precise, general definition of forgetting and answer these questions with the following contributions:
\begin{enumerate}[left=0pt,nosep]
    \item Inspired by \citet{hutter2005,dong2022simple, fortini2019quasi, fong2023martingale}, we define a \emph{general theoretical formulation} for reasoning about how learners acquire, retain, and lose capabilities during learning (\Cref{sec:prelims}).
    \item We \emph{formulate forgetting from a predictive perspective}, defining it as a violation of predictive self-consistency.
    \item\looseness=-1 We define an \emph{operational measure of the propensity to forget} (\Cref{def:forgetfulness}). We then prove that Bayesian updates do not forget (\Cref{thm:bayes_non_forget}) and relate the training of generative models on self-generated data to forgetting (\Cref{sec:retraining}). 
    \item We \emph{empirically study the propensity to forget} in diverse learning paradigms, including regression, classification, generative modelling, CL, and RL. Our results confirm that forgetting dynamics conform to expected characteristics and reveal that forgetting and training efficiency are non-monotonic (\Cref{sec:empirical}).
\end{enumerate}


\section{Related work}
In this paper, we study forgetting as a general property of learning algorithms. While terminology varies across fields, all phenomena involving the loss of previously acquired knowledge reflect the same underlying process. For clarity, we refer to all such phenomena as \emph{forgetting}.

\paragraph{Forgetting in CL.}
Forgetting is often studied in the context of CL \citep{de2021continual, wang2024comprehensive}, where it is typically defined as the loss of performance on previous tasks when new tasks are introduced into training \citep{kirkpatrick2017overcoming, chaudhry2018riemannian, chaudhry2018efficient}.

The literature on task shift distinguishes between two effects: backward transfer, where learning on new data \emph{improves} performance on past tasks \citep{benavides2022theory} in a form of generalisation, and forgetting, where learning on new data \emph{degrades} previously acquired knowledge \citep{mccloskey1989catastrophic, jagielski2022measuring}. 

\looseness=-1
Various approaches have sought to quantify forgetting. In toy settings with known data distributions, forgetting can be measured relative to an oracle \citep{lee2021continual}. Others examine changes to internal representations \citep{kim2025understanding}, model the trade-off between generalisation and forgetting via sequential games \citep{raghavan2021formalizing}, or consider information-theoretic formulations that define forgetting in sequentially learned models as an increase in the effective description length of previously observed data under an updated model \citep{he2020continual}.
Most measures of forgetting are tailored to CL settings within specific domains and cannot be applied across different learning tasks or learning algorithms. Instead, we aim to develop a domain- and learner-agnostic definition and measure of forgetting.

\paragraph{Forgetting in RL.}
Early work in continual RL recognised the risk of forgetting as agents adapt over long horizons \citep{ring1994continual, ring1997child}, and recent surveys show forgetting remains a persistent challenge in RL \citep{khetarpal2022towards}. 

\looseness=-1
\citet{mnih2015human, van2018deep} demonstrate that value-based methods with function approximation often degrade in performance when using earlier estimates. This is closely related to the phenomenon of \textit{policy churn}, where the greedy policy of a value-based learner changes in a large portion of the input space after just a few updates \citep{schaul2022phenomenon}. Furthermore, \citet{ring1994continual, kirkpatrick2017overcoming} demonstrate that policy gradient methods are also prone to overwriting earlier strategies during continual adaptation.
Replay buffers are often used to mitigate these effects by reintroducing past experiences during training \citep{schaul2015prioritized,andrychowicz2017hindsight,random, fedus2020revisiting}.
However, this approach incurs increasing computational and memory overhead as buffer sizes grow and replay frequencies increase \citep{schaul2015prioritized}.

\paragraph{Characterisations of forgetting.}
\looseness=-1
Forgetting is frequently characterised through learner-specific mechanisms.
Model-centric views equate forgetting with \emph{parameter drift} \citep{mccloskey1989catastrophic, french1999catastrophic, kirkpatrick2017overcoming, masse2018alleviating} or, in sequential decision-making settings, \emph{policy drift} \citep{shenfeld2025rl}. By contrast, accuracy-centric views characterise forgetting as \emph{performance decay} on earlier tasks \citep{kemker2018measuring, jagielski2022measuring}.
These mechanism-specific perspectives are useful insofar as the corresponding mechanism is present within the class of learners under consideration.

\looseness=-1
Across research programs (such as continual learning, continual reinforcement learning, domain adaptation, and test-time adaptation), forgetting has been studied in relative isolation, with each field adopting its own assumptions, evaluation methods, and terminology \citep{wang2024comprehensive}.
Despite these differences, each program studies a shared underlying property of a learner: the loss of previously acquired knowledge during a learning update.
This suggests that it would be beneficial to study forgetting as a general property of a learner, independent of the setting.
A general perspective of this kind can improve conceptual clarity and provide a foundation for studying forgetting across disparate research programs.

\looseness=-1
Motivated by this perspective, we study forgetting from a general perspective applicable to any learner, loss, domain, or learning setting. In particular, we treat forgetting as a property of learning updates and develop a theory of forgetting based on the learner's behavioural consistency across updates. We demonstrate the generality of our formalism and show that it enables a measurement of forgetting that empirically corroborates long-standing hypotheses about forgetting.

\section{Learning and inference processes}\label{sec:prelims}
\looseness=-1
We present a general framework in which supervised learning, RL, and generative modelling are all specific cases of a single stochastic interaction process. Our formalism is inspired by the agent-environment perspective in general RL \citep{hutter2005, Lattimore2014, dong2022simple, abel2023convergence, abel2023, hutter2024, kumar2025continual}, adapted to align with conventions in machine learning \citep{bishop2006pattern, goodfellow2016deep, fong2023martingale}.

\paragraph{Notation.}
\looseness=-1
We let capital calligraphic letters denote measurable spaces $(\mathcal X)$\footnote{We assume that all measurable spaces $(\mathcal X,\mathcal Y,\mathcal Z,\dots)$ are standard Borel (Borel $\sigma$-algebras of Polish spaces), such that conditional probability kernels $p(\cdot\mid h,y)$ exist and are measurable.}, lowercase letters denote elements or functions $(f)$, and uppercase italics denote random variables $(X)$. For any measurable space $\mathcal X$, we let $\mathcal{P}(\mathcal X)$ denote the set of probability distributions over $\mathcal X$, where a mapping $p:\mathcal X\times \mathcal Y\to \mathcal{P}(\mathcal Z)$ is interpreted as a family of conditional distributions $p(\cdot\mid x,y)$ on $\mathcal Z$, indexed by $(x,y)\in\mathcal X\times \mathcal Y$.

\subsection{Learning interaction}
We formalise learning as an ongoing interaction between a \emph{learner} and an \emph{environment}, evolving over discrete time steps $t\in\mathbb N_0=\{0,1,\dots\}$.
We distinguish between two sources of stochasticity:
\begin{itemize}[left=0pt,nosep]
    \item \emph{External probabilities} $p_e(\cdot)$ describe the stochasticity inherent to the environment.
    \item \emph{Predictive distributions} $q_f(\cdot)$ describe the predictive uncertainty of the learner.
\end{itemize}
We formalise the data exchanged between a learner and the environment by defining an interface.
\begin{definition}[Interface]
    An environment-learner \emph{interface} is a pair $(\mathcal X,\mathcal Y)$ of measurable spaces.
\end{definition}
\looseness=-1
Elements of $\mathcal X$ are \emph{observations} made by the learner (such as numerical features or targets), and elements of $\mathcal Y$ are \emph{outputs} emitted by the learner (such as actions or predictions).
\begin{definition}[Histories]
    The set of \emph{histories} relative to an interface $(\mathcal X,\mathcal Y)$ is
    \begin{equation}
        \mathcal H=\bigcup_{t\in\mathbb N_0} (\mathcal X\times \mathcal Y)^{t+1}.
    \end{equation}
\end{definition}
An element of a history is a sequence $H_{0:t}=((X_0,Y_0),\dots,(X_t,Y_{t}))\in\mathcal H$ of $t+1$ observation-output pairs.
For a sequence $A=(A_0,\dots,A_t)$ and indices $0\leq i\leq j\leq t$, the subsequence between indices $i$ and $j$ is $A_{i:j}=(A_i,\dots,A_j)$.
\begin{definition}[Environment]
    An \emph{environment} relative to interface $(\mathcal X,\mathcal Y)$ is a pair $(e,p_{X_0})$, where
    \begin{itemize}[left=0pt,nosep]
        \item $e:\mathcal H\times\mathcal Y\to\mathcal{P}(\mathcal X)$ maps each history-output pair to a conditional distribution over the next observation $p_e(\cdot\mid H,Y)$,
        \item $p_{X_0}\in\mathcal{P}(\mathcal X)$ specifies the distribution of the initial observation $X_0$.
    \end{itemize}
\end{definition}
\begin{definition}[Learner]\label{def:learner}
    A \emph{learner} relative to interface $(\mathcal X,\mathcal Y)$ is a tuple $(\mathcal Z,f,u,u',p_{Z_0})$, where:
    \begin{itemize}[left=0pt,nosep]
        \item $\mathcal Z$ is a measurable space called the \emph{learner state space};
        \item $f:\mathcal Z\times\mathcal X\to \mathcal{P}(\mathcal Y)$ is a \emph{prediction function} that returns the conditional distribution $q_f(\cdot\mid z,x)$;
        \item $u:\mathcal Z\times\mathcal X\times\mathcal Y\to \mathcal{P}(\mathcal Z)$ is the \emph{learning-mode state update function};
        \item $u':\mathcal Z\times\mathcal X\times\mathcal Y\to \mathcal{P}(\mathcal Z)$ is the \emph{inference-mode state update function};
        \item  $p_{Z_0}\in\mathcal{P}(\mathcal{Z})$ is the \emph{initial learner state distribution}.
    \end{itemize}
\end{definition}
\looseness=-1
Although a learner's state could, in principle, be updated by a single function (as in work by \citeauthor{dong2022simple}, \citeyear{dong2022simple}, \citeauthor{abel2023convergence}, \citeyear{abel2023convergence}, and \citeauthor{kumar2025continual}, \citeyear{kumar2025continual}), we distinguish between two update functions to capture different modes of learner evolution.
During interaction with the environment, the learning-mode update $u$ governs the evolution of the state, including all components that may affect the learner's predictive distribution.
In contrast, the inference-mode update $u'$ is restricted to state components that do not influence the prediction function $f$: it may update auxiliary components such as memory buffers or counters. Still, it must leave the predictive distribution $q_f$ invariant.
This distinction allows an observer to analyse the learner's behaviour in dynamic settings, where $q_f$ evolves, and in evaluative settings, where $q_f$ is fixed.

\paragraph{The interaction process.}
The interaction between a learner and the environment defines a joint stochastic process.
\begin{definition}[Interaction Process]
    The (learning) \emph{interaction process} between an environment $(e,p_{X_0})$ and a learner $(\mathcal Z,f,u,u',p_{Z_0})$ relative to an interface $(\mathcal X,\mathcal Y)$ is:
    \begin{equation*}
    \begin{aligned}
      (t=0)\text{: } & Y_0=\bot,\quad X_0\sim p_{X_0},\quad Z_0 \sim p_{Z_0},\qquad &&\text{(initialisation)} \\
        (t\ge1)\text{: } & Y_t\sim q_f(\cdot\mid Z_{t-1},X_{t-1}), &&\text{(learner samples output)}\\ 
         & X_t \sim p_e(\cdot\mid H_{0:t-1}, Y_t), &&\text{(environment generates observation)}\\
         & Z_t \sim u(\cdot\mid Z_{t-1},X_t,Y_t). &&\text{(learner updates state)}\\
    \end{aligned}
    \end{equation*}
    This generates the stochastic process $\{X_t,Y_t,Z_t\}_{t\in\mathbb N_0}$, where $Y_0=\bot$ denotes the absence of an output.
    \label{def:interaction}
\end{definition}

\begin{figure}[t]
    \centering
    \begin{subfigure}[t]{0.49\linewidth}
    \begin{tikzpicture}[
        baseline=-1cm,
        state/.style={circle, minimum size=0.65cm, font=\small, fill=violet!20, thick},
        future/.style={rectangle, rounded corners, fill=blue!10, font=\scriptsize, align=center, inner sep=2pt},
        arrow/.style={-{Stealth[inset=0pt, length=5pt, width=4pt]}, thick},
        dist_arrow/.style={-{Stealth[inset=0pt, length=5pt, width=4pt]}, dashed, gray, thick}
    ]
        \node[state] (z0) at (0,0) {$Z_0$};
        \node[state] (z1) at (2.4,0) {$Z_1$};
        \node (dots) at (3.8,0) {\Large $\dots$};
        \node[state] (zt) at (5.2,0) {$Z_t$};

        \draw[arrow] (z0) -- node[pos=0.45, above, font=\scriptsize] {$u$} (z1);
        \draw[arrow] (z1) -- node[pos=0.4, above, font=\scriptsize] {$u$} (3.4,0);
        \draw[arrow] (4.2,0) -- node[pos=0.4, above, font=\scriptsize] {$u$} (zt);
        \draw[arrow] (zt) -- node[pos=0.4, above, font=\scriptsize] {$u$} ++(1.1,0);

        \node[future] (q0_1) at (0, 1.3) {$q(H^{1} \mid Z_0^1, H_0)$};
        \node[future] (q1_1) at (2.4, 1.3) {$q(H^{2} \mid Z_1^2, H_{0:1})$};
        \node[future] (qt_1) at (5.2, 1.3) {$q(H^{t+1} \mid Z_t^{t+1}, H_{0:t})$};

        \draw[dist_arrow] (z0) -- (q0_1);
        \draw[dist_arrow] (z1) -- (q1_1);
        \draw[dist_arrow] (zt) -- (qt_1);

        \node[future] (q0_2) at (0, 2.5) {$q(H^{2} \mid Z_0^2, H_{0:1})$};
        \node[future] (q1_2) at (2.4, 2.5) {$q(H^{3} \mid Z_1^3, H_{0:2})$};
        \node[future] (qt_2) at (5.2, 2.5) {$q(H^{t+2} \mid Z_t^{t+2}, H_{0:t+1})$};

        \draw[dist_arrow] (z0) -- node[pos=0.45, right, black, font=\scriptsize] {$u'$} (q0_1);
        \draw[dist_arrow] (z1) -- node[pos=0.45, right, black, font=\scriptsize] {$u'$} (q1_1);
        \draw[dist_arrow] (zt) -- node[pos=0.45, right, black, font=\scriptsize] {$u'$} (qt_1);
        \draw[dist_arrow] (q0_1) -- node[pos=0.5, right, black, font=\scriptsize] {$u'$} (q0_2);
        \draw[dist_arrow] (q1_1) -- node[pos=0.5, right, black, font=\scriptsize] {$u'$} (q1_2);
        \draw[dist_arrow] (qt_1) -- node[pos=0.5, right, black, font=\scriptsize] {$u'$} (qt_2);

        \node[gray] at (0, 3.1) {$\vdots$};
        \node[gray] at (2.4, 3.1) {$\vdots$};
        \node[gray] at (5.2, 3.1) {$\vdots$};

        \node[gray] at (6.7,0) {$\dots$};
    \end{tikzpicture}
    \end{subfigure}
    \begin{subfigure}[t]{0.49\linewidth}
        \includegraphics[width=\linewidth]{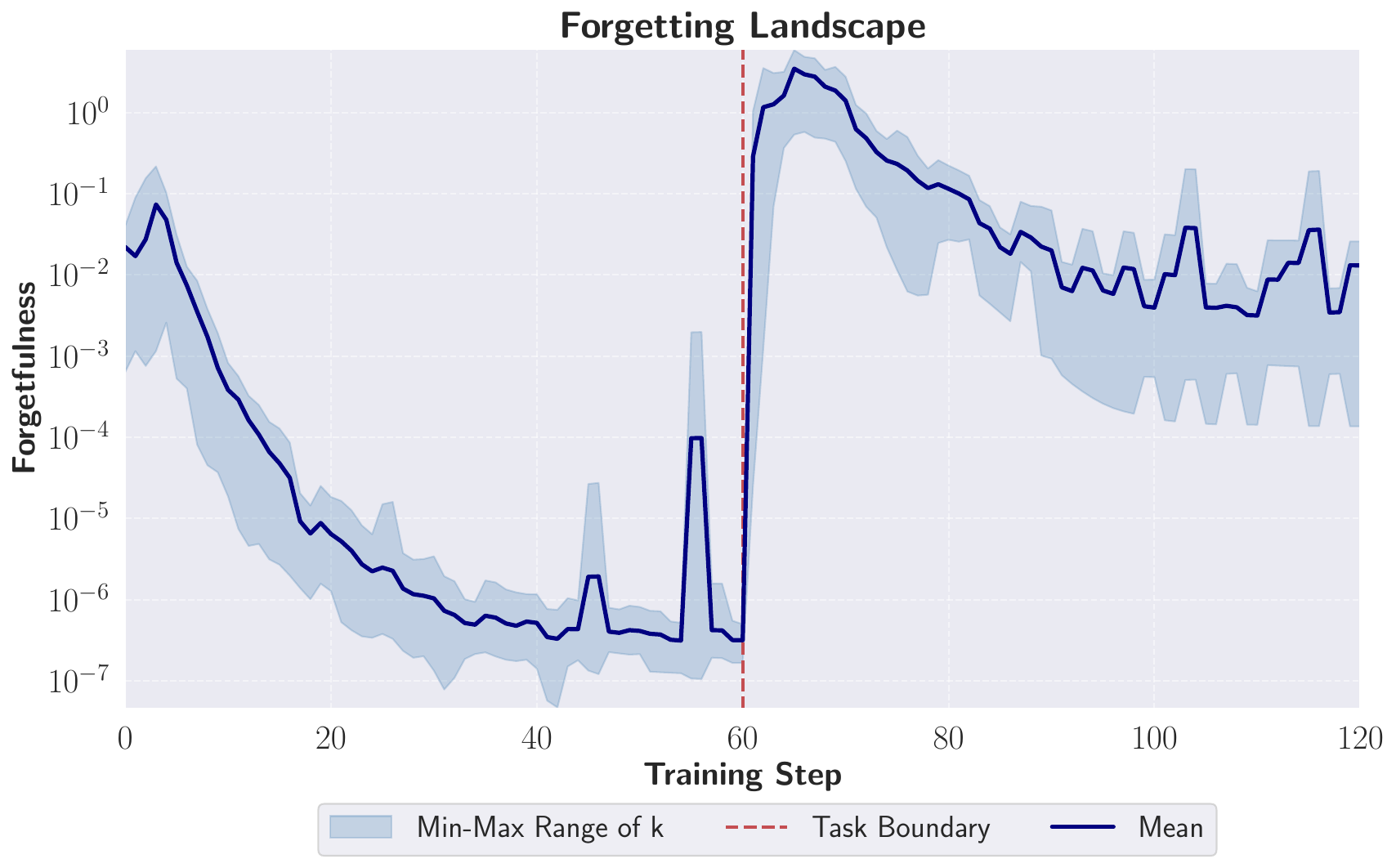}
    \end{subfigure}
    \caption{
    \textbf{State evolution and predictive distributions; forgetting abruptly increases at task boundaries.}
    \textit{Left:} The learner's internal state $Z_t$ evolves through training updates $u$ given $(Z_{t-1},X_t,Y_t)$.
    Each state induces a predictive distribution $q(H^{t+1:\infty}\mid Z_{t-1},H_{0:t})$, which is updated in inference mode $u'$.
    This separation illustrates how each state encodes retained capabilities, and how training vs introspective updates interact. \textit{Right:} The shaded region shows the range, and the solid line shows the mean of the propensity to forget $\Gamma_k(t)$ for $k \in [1, 40]$ of a single-layer neural network trained on a class-incremental two-moons classification task. The red-dashed line marks the task boundary, and results are evaluated across four seeds. This figure demonstrates that our conceptual framework clarifies when and why forgetting occurs, confirming prior intuition that forgetting increases sharply at task boundaries, while providing a detailed view of the forgetting profile across the entire learning process. Implementation details are provided in \Cref{app:implementation}.}
    \label{fig:conditional_futures}
    \vspace{-1em}
\end{figure}
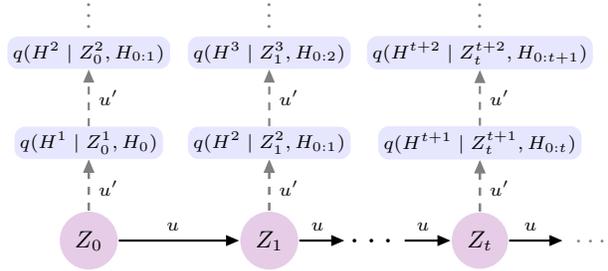
\looseness=-1
An important result in CL theory is that optimal CL is NP-hard, as demonstrated by \citet{CL-NPhard}.
This motivates the need for theories that enable explicit reasoning about bounded learners.
The learner-state perspective provides such a foundation by modelling learning as the evolution of a finite agent state that mediates all behaviour, as demonstrated by \citet{abel2023convergence}.
In this view, the space of admissible state updates is constrained by finite computational and memory resources, and forgetting arises through the update rule $u$.

\subsection{Predictive distributions}\label{sec:futures}
During learning, the learner's state encodes expectations about how future interactions will unfold. We can make these expectations explicit by asking the learner to \emph{internally simulate the future} using update $u'$. This creates a distribution over an infinite sequence of future interactions.

\looseness=-1
This simulation yields a predictive distribution that represents a hypothetical rollout of the learner's predictive model under its current beliefs.
This is \emph{entirely isolated from the interaction process}.
To distinguish this introspective evolution from real learning progression, we introduce introspection time indices $S\in\mathbb N$ placed as superscripts on $X$, $Y$, and $Z$. 

During the rollout, the learner generates targets from its own predictive distribution.
Formally, given a history $H_{0:t}$ and a state $Z_t$, the predictive distribution evolves according to
\begin{equation*}\label{eq:inference_update}
    X^s\sim q_e(\cdot\mid Y^s,Z_t^{s-1})\qquad Y^s\sim q_f(\cdot\mid Z_t^{s-1},X^{s-1})\qquad Z_t^s\sim u'(Z_t^{s-1},X^{s},Y^s),
\end{equation*}
\looseness=-1
where $s\in S=\{t+1,t+2,\dots\}$.
Here, $q_e$ denotes the observation-generation kernel used during rollout; it coincides with the learner's generative model for any components of $X$ that are explicitly modelled, and samples unmodelled components (e.g., inputs to a discriminative model) from an external data source such as a validation set.
Formally, $q_e$ may be a product kernel combining learner-modelled and externally supplied marginals.
An element of future histories $H\in\mathcal H$ is a sequence of future observation-output pairs $(X^s,Y^s)$ induced by the learner.
\begin{definition}[Predictive Distributions]
    The \emph{predictive distribution} of the learner at state $Z_{t}$ with realised history $H_{0:t}$ is the joint distribution obtained by simulating the inference-mode process in \eqref{eq:inference_update}:
    \begin{equation}
        q(H^{t+1:\infty}\mid Z_{t},H_{0:t})
    \end{equation}
\end{definition}
\looseness=-1
Thus, the predictive distribution is a probability distribution over the space of infinite sequences of future inputs and outputs, $(\mathcal{X}\times\mathcal{Y})^{\mathbb{N}}$.
As the learner's state $Z_t$ and future are stochastic, the predictive distribution at each step $q(\cdot\mid Z_t, H_{0:t})$ is a random variable taking values in $\mathcal{P}((\mathcal{X}\times\mathcal{Y})^{\mathbb{N}})$.

In practice, the predictive distribution is given by the learner's output parameterisation.
For classification, this corresponds to a categorical (or Bernoulli) distribution over classes; for regression, to the probability of the target specified by the learned model (typically Gaussian under standard losses); and for generative models, to the induced distribution over data samples defined by the generative process.
For a detailed discussion, see \Cref{app:empirical}.

\paragraph{Predictive-Bayesian perspective.}
\looseness=-1
This construction is inspired by predictive Bayesianism \citep{fortini2019quasi, fortini2025exchangeability} and the martingale posterior framework \citep{fong2023martingale}.
Under this view, each learner implicitly maintains an infinite sequence of future observations, and the resulting statistics form a martingale that converges.
A central challenge in CL is that learning occurs in the parameter space, whereas the quantities of interest exist in the data space.
Predictive Bayes provides one plausible way to relate parameter updates to predictive statements.
This allows predictive statements to be \emph{validated} against realised outcomes and provides a representation of the learner's knowledge.

\subsection{Learning as a stochastic process}\label{sec:learning_process}
Throughout the interaction process, the sequence $\{Z_t\}$ generates a stochastic process over predictive distributions.
Thus, the distribution over learner states at time $t$ defines a distribution over predictive distributions (illustrated in \Cref{fig:conditional_futures}).

\paragraph{Interpretation across learning paradigms.}
\looseness=-1
The variables $(X_t, Y_t, Z_t)$ admit natural interpretations across learning settings, summarised in \Cref{tab:framework}. In supervised learning, $X_t$ denotes (current input, previous target) pairs and $Y_t$ is the learner's output in response to $X_{t-1}$; in RL, $X_t$ contains observations and rewards while $Y_t$ are actions. Across these paradigms, the learner state $Z_t$ encapsulates all the learner's contents (including parameters, buffers, etc.).

\looseness=-1
For example, in a supervised setting, $X_t$ contains the input $x_t$ and the target $y_{t-1}$ for the previous input. The learner's output $Y_t$ is conditionally generated from the previous observation $X_{t-1}$ and is interpreted as a prediction on $x_{t-1}$. The environment then produces $X_t$, revealing the true label $y_{t-1}$ associated with prediction $Y_t$ and presents the next input $x_t$. The learner state $Z_t$ is updated in learning-mode via $u$, and the inference-mode update $u'$ can be used to derive the predictive distribution of the state $Z_t$ at any time. Both $u$ and $u'$ are conditioned upon $(Z_{t-1}, X_t,Y_t)$.

\looseness=-1
While the same symbols are used throughout, their interpretation shifts depending on the paradigm. What remains consistent is the structure of the interaction process. This allows us to represent general learning processes with a single stochastic-process formalism that details the evolution of the inputs $X_t$, outputs $Y_t$, and the learner's state $Z_t$.

\section{Forgetting}
Upon receiving a new observation, the learner updates its state and then uses that updated state to produce an output.
Therefore, learning updates constitute a dual process: the acquisition of new information from observation and the loss of information in the previous state.
We isolate forgetting as the latter phenomenon---the loss of information from the prior state induced by an update--- thereby distinguishing it from changes attributable to newly acquired information.
This perspective leads to a definition of forgetting as the extent to which an update induces behavioural inconsistency.

\subsection{Characterising forgetting}
Before formalising forgetting, we first outline desiderata that any valid notion of forgetting should satisfy. These desiderata are motivated by the thought experiments in \Cref{app:thought_exps}.
\begin{desideratum}\label{des:1}
    A forgetting measure should quantify the loss of learned information over time.
\end{desideratum}
\looseness=-1
Forgetting is distinct from measurable success on a task, such as accuracy or cumulative reward. It is a property of the learner, independent of the environment. A learner can maintain outdated or incorrect beliefs yet still perform well, or lose relevant knowledge without an immediate change in performance. Conventional metrics, including backward transfer, may fail to capture these underlying dynamics, highlighting the need to disentangle forgetting from performance.
\begin{desideratum}\label{des:2}
    A characterisation of forgetting must not conflate forgetting with the correctness of outputs or with justified updates that change beliefs.
\end{desideratum}
When a learner incorporates new observations, their beliefs (and state) will change. \emph{A change in belief does not necessarily imply that anything is forgotten.} Therefore, conceptualisations based on changes to beliefs (or parameters) can misidentify information-preserving updates as forgetting. 
\begin{desideratum}\label{des:3}
    Forgetting should characterise the learner's loss of prior information, not just the retention of previously observed data.
\end{desideratum}
Forgetting encompasses the loss of information, not only individual observations. For example, a learner may forget how to generalise to unobserved examples that they were previously able to process. Conceptualisations of forgetting that prioritise memorisation overlook this broader notion.
\begin{desideratum}\label{des:4}
    Forgetting is a property of the learner, not of the environment in which it operates.
\end{desideratum}
An environment cannot forget; however, it can influence the rate or magnitude of forgetting.
The thought experiments in \Cref{app:thought_exps} justify the desiderata and are the foundation of the forgetting formalism developed below.

\subsection{Forgetting and self-consistency}\label{sec:forgetting}
\looseness=-1
We identify a condition that implies a learner does not forget.
Before an update, the learner's state induces a predictive distribution that represents the learner's current behavioural dispositions.
Given a new input, the learning update induces a revised predictive distribution that reflects the learner's new behavioural dispositions.
A natural requirement for non-forgetting is that learning should preserve behaviour associated with information that the learner has already encoded, and that is not contradicted or extended by the new input.
Thus, we can consider the case in which the learner is presented with inputs that contain no new information relative to its prior state.
In this case, a non-forgetting learner should remain unchanged in its behaviour.
This requirement is a behavioural consistency condition.

Therefore, we define forgetting in terms of changes to the learner's predictive distribution.
Given that the learner's predictive distribution contains information the learner has already encoded, updating it with such introspective data introduces no new information and should not change the learner's behaviour.
If, on average, these updates induce changes in the learner's behaviour, this indicates that previously acquired information has not been preserved through the update, implying that the update is forgetful.

\looseness=-1
The introspective data is constructed by marginalising over $k\in\mathbb N$ updates applied to targets sampled from the learner.
\begin{definition}[Simulated Marginalisation]\label{def:sim_marg}
    The $k$-step \emph{simulated marginalisation} $q_k^*(H^{t+k:\infty}\mid Z_{t-1},H_{0:t-1})$ induced by updating on samples from the learner's predictive distribution is given by
    \begin{equation}
        q_k^*(H^{t+k:\infty}\mid Z_{t-1},H_{0:t-1})\coloneqq \mathbb{E}_{X_{t:t'},Y_{t:t'},Z_{t:t'}}\left[ q(H^{t+k:\infty}\mid Z_{t'},H_{0:t'})\right],
    \end{equation}
    where $k\in\mathbb N$, $t' = t+k-1$, and for $i=t, \ldots, t'$ the expectation is taken over $X_i\sim q_e(\cdot\mid Y_i, Z_{i-1})$, $Y_i\sim q_f(\cdot\mid Z_{i-1},X_{i-1})$, $Z_i\sim u(\cdot\mid Z_{i-1},X_i,Y_i)$.
\end{definition}
In this formulation, the expectation accounts for the stochasticity of the learner's inputs, actions, and transitions.
A learner does not forget if the predictive distribution before and after the simulated marginalisation is equivalent.
This yields the following notion of predictive self-consistency.
\begin{definition}[Self-Consistency Condition]
    \label{def:consistency_condition}
    A learner is $k$-step \emph{self-consistent} if and only if
    \begin{equation}
        q(H^{t+k:\infty} \mid Z_{t-1}, H_{0:t-1}) = q_k^*(H^{t+k:\infty} \mid Z_{t-1}, H_{0:t-1}).
    \end{equation}
\end{definition}

\paragraph{Propensity to forget.}\label{sec:measure} 
In practice, most learning algorithms do forget. To quantify \emph{how much} the learner is likely to forget at any time, we introduce a measure grounded in our formalism. When the self-consistency condition (\Cref{def:consistency_condition}) is violated, the learner's predictive distribution after the simulated marginalisation diverges from the initial predictive distribution. Measuring this divergence yields a notion of the learner's \emph{propensity to forget}, providing an algorithm- and task-agnostic measure of forgetting.
\begin{definition}[Propensity to Forget]
\label{def:forgetfulness}
    The $k$-step \emph{propensity to forget} incurred at time $t$ is given by
    \begin{equation}
        \Gamma_k(t) \coloneqq \mathrm{D}\big( q(H^{t+k:\infty}\mid Z_{t-1},H_{0:t-1})\,\| \, q_k^*(H^{t+k:\infty}\mid Z_{t-1},H_{0:t-1}) \big).
    \end{equation}
    where $\mathrm D(\cdot\|\cdot)$ is a suitable divergence measure.
\end{definition}


\paragraph{Scope of theory.}
\looseness=-1
Our formalism applies to all components of the learner's state that influence the predictive distribution.
For instance, neural network parameters or replay-buffer entries that are later sampled during training influence behaviour and are therefore represented by the predictive distribution.
By contrast, state components that never affect behaviour are outside the scope of the formalism.
For instance, unused memory allocations or replay-buffer entries that are never sampled can be changed or deleted without altering the learner's behaviour; therefore, these components are not captured by the predictive distribution, and their loss is not captured by the theory.
Accordingly, this formalism characterises forgetting in terms of behaviour induced by learning updates, while excluding changes to state that never influence behaviour.

\section{Analysis}\label{sec:empirical}
\begin{figure*}[t]
    \includegraphics[width=0.8\linewidth]{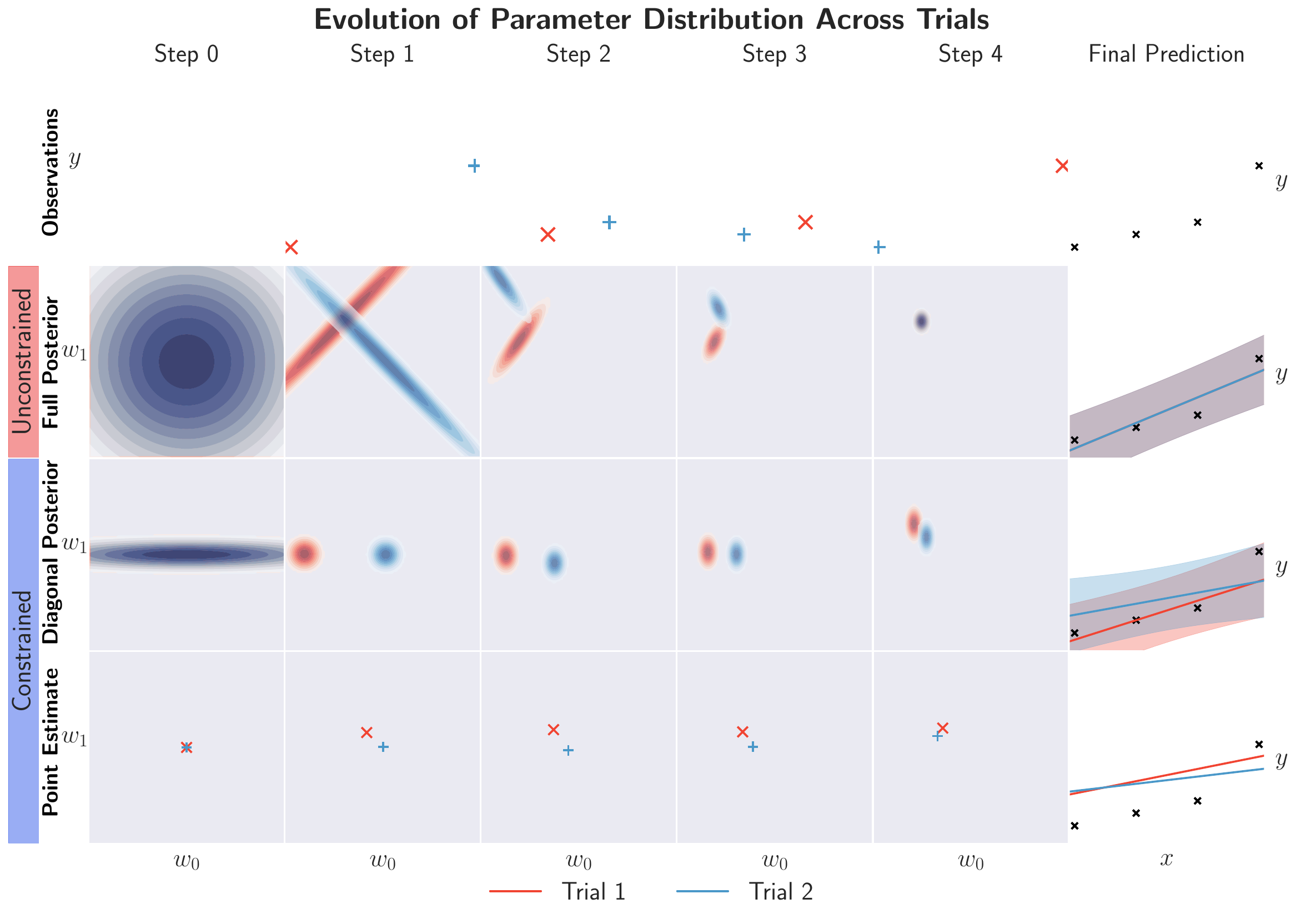}
    \centering
    \caption{\textbf{Exact Bayesian learners can adapt without forgetting.}
      Axes showing observations $(x, y)$ are shaded white, and the parameter axes $(w_0,w_1)$ are shaded grey.
      \textit{Top row:} The same four observations are presented to a linear regression learner in different orders.
      \textit{Final column:} The resulting posterior predictive distributions.
      \textit{Second row:} An exact Bayesian posterior can update its parameters without forgetting.
      \textit{Bottom rows:} Two constrained learners -- a Gaussian variational posterior with diagonal covariance updated by variational Bayes and a point estimate updated by gradient descent -- are not permutation-invariant in exchangeable settings and violate the self-consistency condition.
    }
    \label{fig:exchangeable}
    \vspace{-1em}
\end{figure*}
\Cref{def:forgetfulness} establishes a precise object of study for the analysis of learning dynamics.
In this section, we examine the consequences of forgetting through both empirical and theoretical analysis.
Theoretically, we show that exact Bayesian updates are not forgetful.
Empirically, we demonstrate that gradient descent induces forgetting across standard learning paradigms, including supervised, continual, and reinforcement learning.
Importantly, we show that forgetting is influenced by design choices such as the update rule, architecture, and hyperparameters, each of which induces distinct forgetting profiles that affect learning dynamics and efficiency.
These results are obtained in simple, widely used settings, demonstrating that forgetting is a fundamental property of learning systems, arising even in the absence of task structure or distributional shift.
We use these results to develop a more intuitive and unified account of forgetting.

\subsection{Unforgetful learners}\label{sec:unforgetful}
We first confirm an intuition about our definition of forgetting: an exact Bayesian learner does not forget, even if its model is misspecified. We formalise the notion of such a learner and present the proof of the theorem in \Cref{app:bayesian_theorem}.

\begin{theorem}[Bayesian Non-Forgetting]\label{thm:bayes_non_forget}
    Let $k\in\mathbb N$ and for every $t\in\mathbb N$ define $t'=t+k-1$. Under \Cref{assumption:updates,assumption:markov,assumption:mixture,assumption:likelihood,assumption:regularity}, the learner's posterior-predictive distribution satisfies
    \begin{equation}
        q(H^{t+k:\infty} \mid Z_{t-1}, H_{0:t-1})= q_{k}^*(H^{t+k:\infty} \mid Z_{t-1}, H_{0:t-1}),
    \end{equation}
    where $Z_t$ is derived by deterministic Bayesian updates.
    Thus, $\Gamma_k(t)=0$ for all $k,t\in\mathbb N$.
\end{theorem}
\looseness=-1
Filtrations formalise the accumulation of information over time in non-forgetting settings.
In this perspective, the learner's predictive distribution can be seen as a stochastic process adapted to a state filtration.
This represents all information the learner has encoded up to time $t$.
Exact Bayesian updates ensure that this stochastic process forms a martingale with respect to the state filtration \citep{fong2023martingale}.
This indicates that, in expectation, updating the learner state on a simulated future observation does not change the predictive distribution.
Violations of this property correspond to a drift in the predictive distribution, manifesting as forgetting.
For a detailed discussion, see \Cref{app:martingale}.

\looseness=-1
Under this predictive notion of forgetting, exact Bayesian learners exhibit no forgetting.
The only other update rule we identify with this property is the identity update, which ignores new data entirely. This motivates the following question.
\begin{openquestion}
    Are exact Bayesian updates the unique non-degenerate learning update rule that avoids forgetting?
\end{openquestion}
In contrast, \emph{approximate} Bayesian learners, such as those based on variational approximations, can exhibit forgetting (\Cref{fig:exchangeable}). Investigating the forgetting behaviour of existing approximate posterior update approaches and identifying approaches that improve them are valuable directions for future work.

\vspace{-0.75em}
\paragraph{Parameter updates.}
\looseness=-1
A prevalent idea in continual learning is that parameter updates necessarily cause forgetting \citep{mccloskey1989catastrophic, french1999catastrophic, kirkpatrick2017overcoming, zenke2017continual, aljundi2018memory, masse2018alleviating, li2024spirf, zhao2023does, shenfeld2025rl}.
This has led to the development of regularisation techniques that attempt to mitigate forgetting by penalising parameter changes \citep{kirkpatrick2017overcoming,zenke2017continual}.

\looseness=-1
However, this perspective conflates changes to parameters with information loss.
Exact Bayesian learners provide a counterexample.
Although their parameters are updated with each new observation, the resulting posterior can remain self-consistent (\Cref{fig:exchangeable}).
Thus, Bayesian learners can change their parameters without forgetting.
\begin{takeaway}
    \emph{Bayesian updates ensure self-consistent predictive distributions over time. This self-consistency guarantees that exact Bayesian updates do not forget information, even as they adapt to new observations.}
\end{takeaway}

\begin{figure*}[t]
    \centering
    \begin{subfigure}[b]{0.33\linewidth}
        \includegraphics[width=\linewidth]{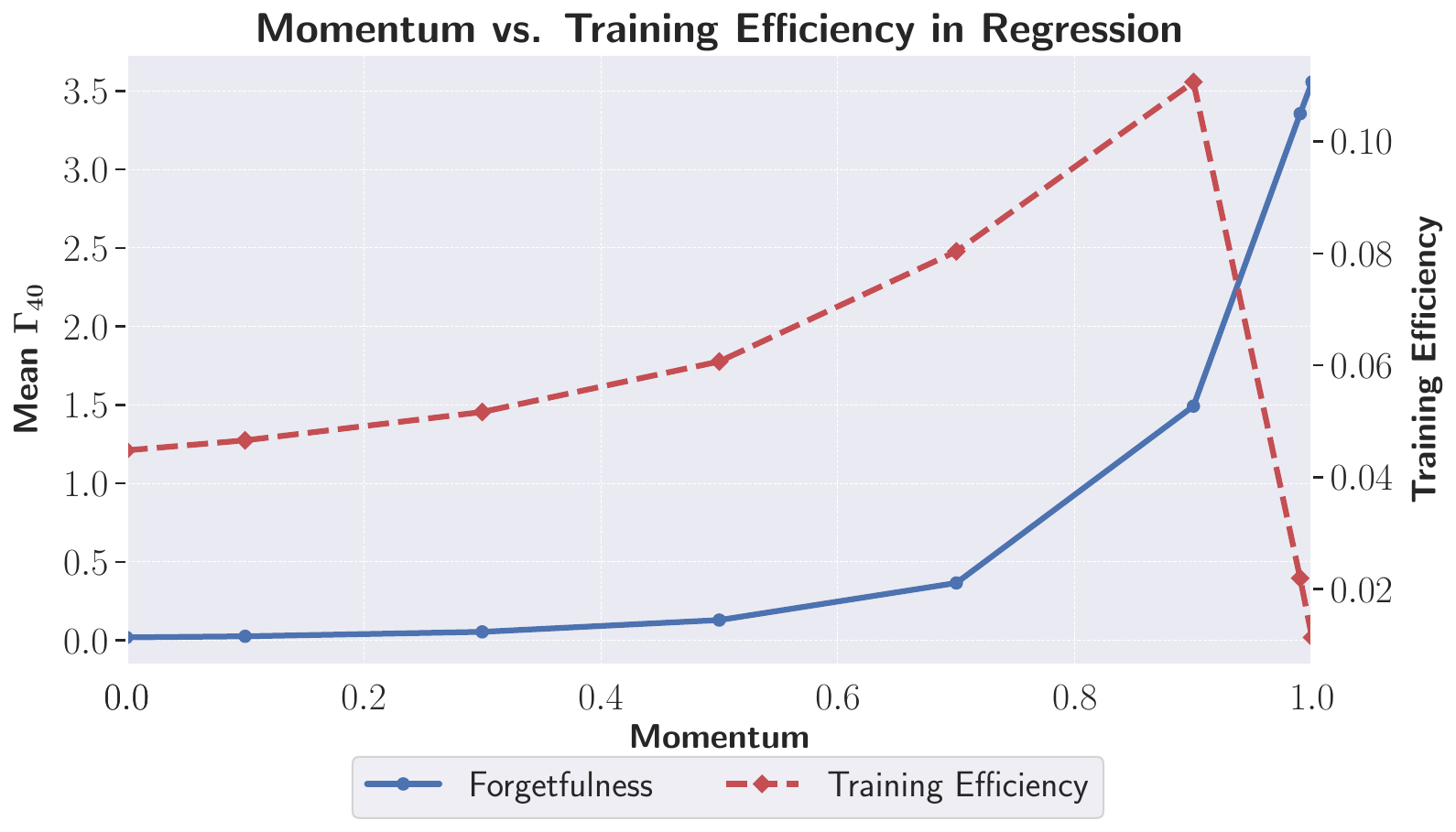}
    \end{subfigure}
    \begin{subfigure}[b]{0.33\linewidth}
        \includegraphics[width=\linewidth]{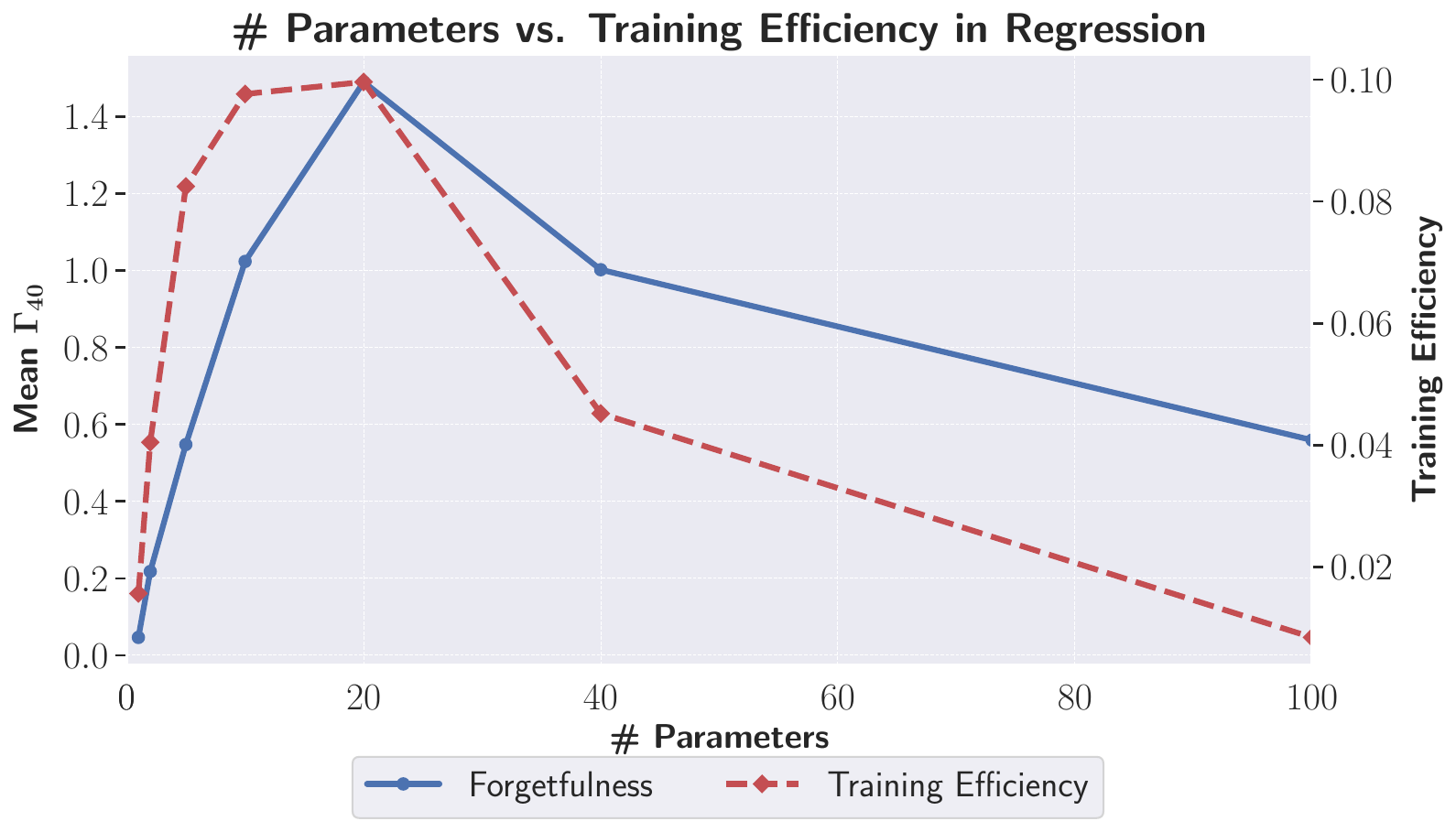}
    \end{subfigure}
    \begin{subfigure}[t]{0.33\linewidth}
        \includegraphics[width=\linewidth]{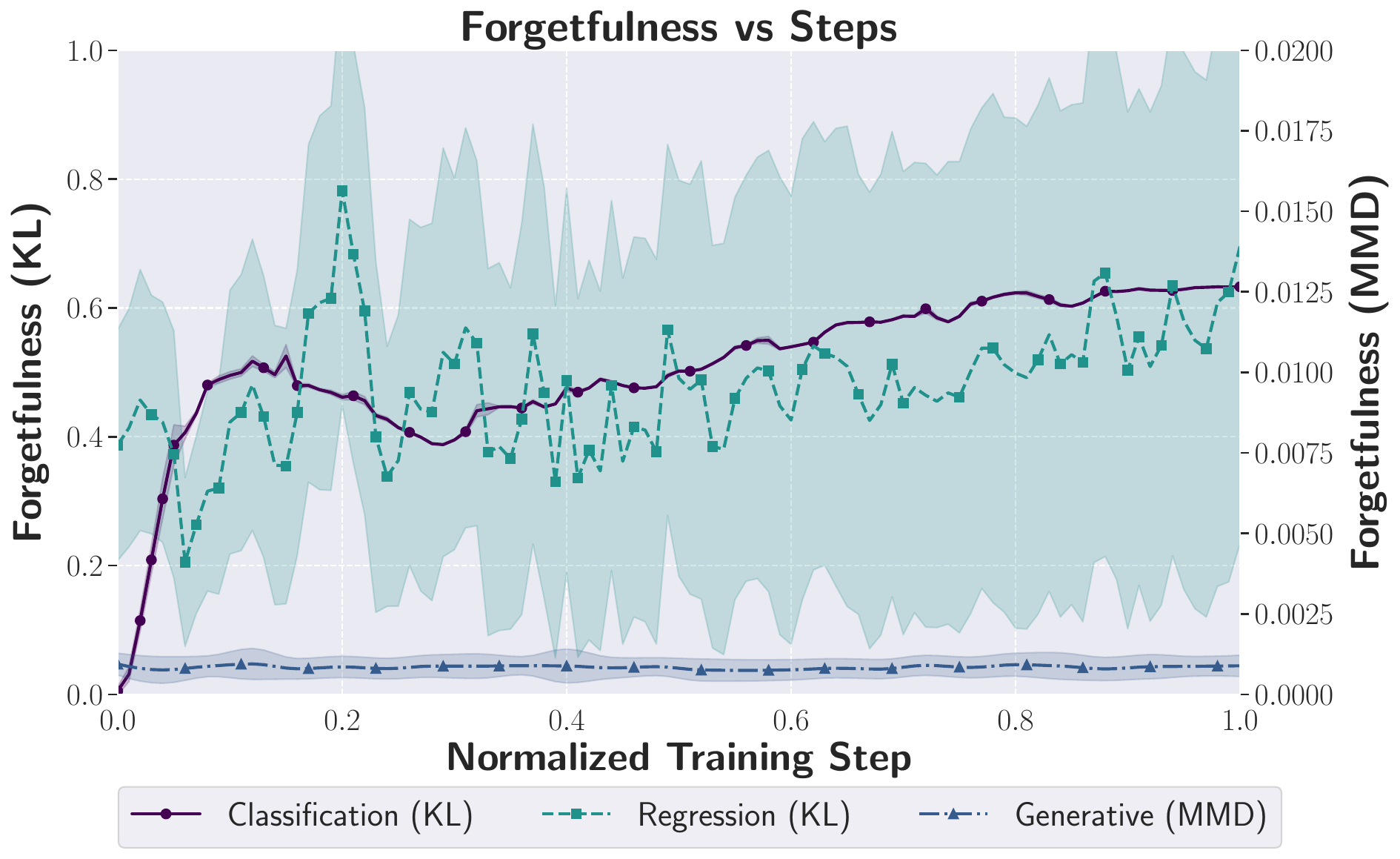}
    \end{subfigure}
  \caption{\looseness=-1\textbf{Approximate learners can benefit from forgetting; forgetting occurs across all deep learning scenarios.}  \emph{Left/Middle:} Training efficiency and forgetfulness across varying momentum, $\mu$, (left) and number of parameters, $m$, (middle) for a regression task. Efficiency is proxied by the inverse of the normalised area under the training loss curve; forgetting is the mean $k$-step propensity to forget, $\Gamma_{40}(t)$, averaged over training. We observe that maximum training efficiency does not coincide with minimal forgetting, as insufficient forgetting slows adaptation while excessive forgetting destabilises learning. \emph{Right:} Forgetfulness dynamics of a shallow neural network trained on regression, classification, and generative modelling tasks. Regression and classification tasks use KL divergence, while the generative task uses the maximum mean discrepancy (MMD). We show the mean $k$-step forgetfulness ($k\in[1,40]$) over the normalised training step, with the shaded region indicating the range. Forgetting dynamics vary without any distribution shift. We observe similar dynamics in the high-dimensional experiment conducted in \Cref{app:forgetting_supervised} and \Cref{fig:cifar-forgetfulness}. Implementation details of the classification, generation, and regression experiments are provided in \Cref{app:implementation}.}
    \label{fig:training_efficiency_il_forgetting}
    \vspace{-1em}
\end{figure*}

\subsection{Forgetting in deep learning}\label{sec:diff_settings}
Across deep learning settings, forgetting is nonzero, with both its magnitude and dynamics varying across settings (\Cref{fig:training_efficiency_il_forgetting}).
Even in i.i.d.\ settings, forgetting fluctuates throughout training because each gradient-descent step updates the learner's parameters, thereby altering the learner's state and inducing changes to the predictive distribution.
Since gradient descent updates do not satisfy the self-consistency condition in \Cref{def:consistency_condition}, the updates are forgetful even when the data distribution is stationary.
Although absolute $\Gamma_k(t)$ values are domain-specific, the forgetting trajectory reveals how the learner's knowledge evolves.
Therefore, forgetting is functionally meaningful in all tasks, highlighting the importance of forgetting and motivating our paper title, ``Forgetting is Everywhere".

\vspace{-0.75em}
\paragraph{Approximate learners can benefit from forgetting.}
\looseness=-1
At each update, an approximation-based learner incorporates new information from current observations while discarding parts of the existing information in its state (\Cref{sec:forgetting}). Therefore, performance depends on striking a balance between adapting to new information and retaining useful prior information. This is known as the stability-plasticity trade-off \citep{kirkpatrick2017overcoming,parisi2019continual,dohare2023maintaining}.

\looseness=-1
To study this effect, we investigate how modifications to the learner influence the propensity to forget. Throughout, we observe \emph{a moderate amount of forgetting improves learning efficiency}. We quantify training efficiency using the inverse of the normalised area under the training loss curve, a practical optimisation proxy for the rate of convergence. The forgetting-efficiency relationship exhibits an ``elbow" (\Cref{fig:training_efficiency_il_forgetting}), indicating that effective approximate learners utilise forgetting as a mechanism for efficient adaptation. While often discussed in non-stationary settings, \Cref{fig:training_efficiency_il_forgetting} demonstrates that the stability-plasticity trade-off is always present in neural networks, even in stationary tasks. \Cref{app:results} presents ablation studies demonstrating non-monotonicity between forgetting and training efficiency in deep learning across model architectures, high-dimensional datasets, and hyperparameters.
\vspace{-0.15em}
\begin{takeaway}
    \emph{For approximate learners, continual integration of new information requires selectively discarding current knowledge while consolidating new knowledge. Therefore, they must utilise forgetting to adapt over time.}
\end{takeaway}
\vspace{-0.25em}

\subsection{Training on self-generated data}\label{sec:retraining}
\looseness=-1
Generative models often degrade when trained on their own outputs \citep{alemohammad2023self,bertrand2023stability}.
Consequently, generated outputs become less representative of real-world data.
This presents a significant challenge for scaling training pipelines that rely on synthetic (machine-generated) content.

\looseness=-1
While prior work has linked this effect to forgetting \citep{shumailov2023curse,scholten2025model}, a detailed explanation for why degradation occurs remains elusive.
To address this gap, we present a tautological explanation for this phenomenon.
Examining the retraining process reveals that retraining on synthetic data is analogous to a simulated marginalisation (\Cref{def:sim_marg}).
During retraining, this marginalisation occurs explicitly and directly affects the predictive distribution.
Under \Cref{def:consistency_condition}, any change in the predictive distribution when training on such self-generated data constitutes forgetting by definition, explaining the inevitability of degradation.

\looseness=-1
Our theory provides a new perspective on the broad range of phenomena that are influenced by forgetting.
Further instances we explore include challenges faced by RL algorithms (see \Cref{sec:dist_shift}) and reasons for improvements achieved in CL via replay (see \Cref{app:replay-theory}).
These examples are not exhaustive; we believe that many additional phenomena linked to forgetting remain to be explored.
Taken together, these observations demonstrate that the influence of forgetting is widespread across learning systems, demonstrating the need for a general, abstract understanding of forgetting.

\subsection{Implications of distribution shift}\label{sec:dist_shift}
\looseness=-1
Distribution shift and stochasticity influence forgetting dynamics.
In i.i.d.\ settings, the interaction process is stationary, and the predictive distributions are stable.
When hyperparameters are well-tuned, the learner effectively balances adaptation and retention.
These conditions yield desirable self-consistency properties that improve learning efficiency.
\begin{figure*}[t]
    \centering
    \includegraphics[width=0.95\linewidth]{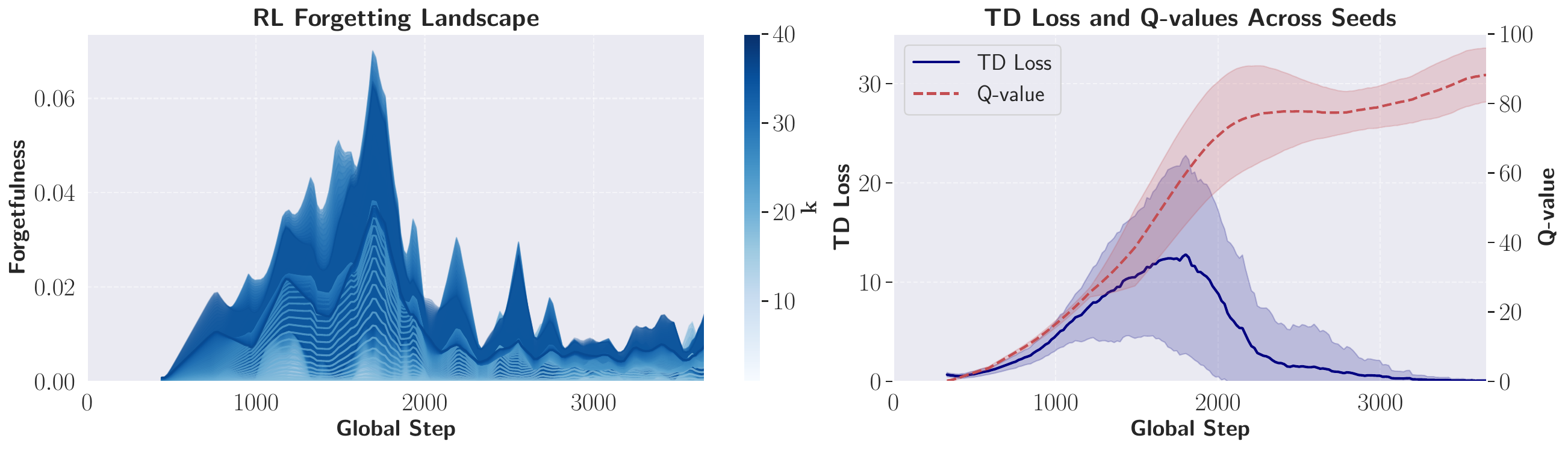}
    \caption{\looseness=-1
    \textbf{DQN actively manages the information acquisition-retention trade-off.}
      We show TD loss, Q-value evaluation, and the forgetting profile of a DQN learner trained on cartpole across ten seeds.
      Early in training, TD loss is low, rises as the agent acquires new information, and then decreases once knowledge has been consolidated.
    The forgetting curve follows this trajectory, suggesting forgetting is a mechanism for balancing knowledge acquisition and retention.
    }
    \label{fig:rl_forgetting_landscape}
    \vspace{-1em}
\end{figure*}

However, in CL, abrupt changes in the observation distribution can cause abrupt changes in the state $Z_t$. Consequently, the magnitude of self-consistency \emph{violation abruptly increases at task boundaries} (\Cref{fig:conditional_futures}).

Learning involves balancing the integration of new information with the retention of current knowledge.
RL presents this challenge in an extreme form; the learner's policy influences future observations, inducing continual non-stationarity.
In DQN, the TD loss rises as the agent incorporates new information.
As the agent consolidates this information, the TD loss and the rate of information acquisition decline (\Cref{fig:rl_forgetting_landscape}).
The forgetting curve tracks TD loss because forgetting is the mechanism the agent utilises to manage this process, demonstrating its importance in RL.

\begin{takeaway}
    \emph{In non-stationary environments, continual changes to the data distribution often simultaneously outpace a neural network's ability to adapt and exacerbate its predictive self-inconsistency. As a result, the neural network struggles to consolidate information and select the appropriate information to discard.}
\end{takeaway}

\subsection{Interpreting prior notions of forgetting}
An advantage of formal theories is that they provide conceptual clarity, helping us formalise intuition and clarify what we mean when discussing certain concepts.
We demonstrate how our theory achieves such clarity by jointly analysing measures of forgetting and methods that mitigate forgetting within our general theory.

\vspace{-0.5em}
\paragraph{Measures of forgetting.}
Existing measures can be understood as differing along two axes: the distribution over which behavioural change is evaluated, and whether forgetting is measured directly in behaviour or indirectly through task-level performance.
Throughout the literature, forgetting is typically estimated under restricted evaluation distributions.

Example-level approaches \citep{toneva2018empirical} evaluate consistency over the training support, task-based methods \citep{kirkpatrick2017overcoming,chaudhry2018riemannian,kemker2018measuring} aggregate performance across predefined task partitions, and KL-based analyses of fine-tuning dynamics \citep{shenfeld2025rl} measure changes in behaviour under adaptation distributions.
In each case, forgetting is estimated relative to a particular evaluation support rather than across the learner's full behavioural domain.

These approaches also differ in how behaviour is represented.
Some methods compare behaviour directly \citep{shenfeld2025rl}, while others first map behaviour to a task-specific performance metric such as accuracy \citep{kirkpatrick2017overcoming,toneva2018empirical,chaudhry2018efficient,kemker2018measuring}.
This mapping is generally non-invertible and lossy, as changes in behaviour can occur without changing performance on the task.
Consequently, performance-based measures provide a partial view of behavioural change, although they are often easier to estimate in practice.
However, none of these approaches directly measures the behavioural consistency of the update rule.
Therefore, they often capture multiple interacting properties of learning dynamics simultaneously, such as backward transfer.

\vspace{-0.5em}
\paragraph{Methods that mitigate forgetting.}
The same perspective also provides a basis for reasoning about methods designed to mitigate forgetting.
Consider Fisher-information-based regularisation methods, such as \textit{elastic weight consolidation} (EWC) \citep{kirkpatrick2017overcoming}.
The Fisher information approximates the sensitivity of the learner's behaviour to parameter perturbations under a reference distribution, and EWC penalises updates in directions that are highly sensitive to parameter perturbations.
From this perspective, EWC discourages updates that are expected to substantially alter behaviour on the reference distribution.
However, constraining parameter sensitivity does not necessarily guarantee the preservation of global behaviour after learning updates.
Moreover, behavioural consistency can sometimes be preserved despite substantial changes in the underlying predictive distribution, as discussed in \Cref{sec:unforgetful} and demonstrated in \Cref{fig:exchangeable}.

\section{Conclusion}
\looseness=-1
In this work, we proposed an algorithm- and task-agnostic formulation of forgetting, characterising it as the \emph{self-inconsistency of a learner's predictive distribution}.
This provides a unified theory of forgetting that admits an operational measure (\Cref{def:forgetfulness}), and is independent of backward transfer.

Using our formalism, we demonstrated its theoretical utility by providing a tautological explanation for why generative models forget when retrained on their own data. We also prove that Bayesian learners do not forget (\Cref{thm:bayes_non_forget}).
This formalises the intuition that some learners adapt to new data without forgetting and demonstrates the widespread applicability of our theory to understanding forgetting.

Our empirical analysis (\Cref{sec:diff_settings}) shows that forgetting is ubiquitous in deep learning and is influenced by interactions between the learner and the environment.
We observe that training efficiency and forgetting are not monotonically related: in the settings we study, an intermediate amount of forgetting can maximise learning efficiency.
Collectively, these findings demonstrate the importance of considering forgetting when designing and evaluating learning algorithms.

Overall, our work introduces a perspective on forgetting that is widely applicable across algorithms and provides novel insights into why algorithms forget.
We hope our work establishes a foundation for analysing how learning algorithms acquire, maintain, and lose capabilities over time, guiding the development of new machine learning algorithms.

\paragraph{Open questions.}
\looseness=-1
This work opens several directions for future research.
Our theory provides a unified perspective on forgetting, suggesting that a natural next step is to apply it across research programs in which it has traditionally been studied in isolation.
Within the theory, we isolate forgetting as the loss of information from the previous state during an update.
A complementary notion is the failure to fully incorporate information from the current observation, and understanding the interaction between these two effects remains an interesting open problem.
Another direction concerns the characterisation of non-forgetful learning algorithms.
While we show that exact Bayesian updates satisfy this property, a natural question is whether other classes of update rules can achieve the same behaviour.
Finally, an important avenue for future work is to understand how boundedness and approximation influence forgetting.

\subsubsection*{Acknowledgements}
\label{sec:ack}
The authors would like to thank Guy Frankel, Jonathon Hare, Henry Gouk, Andras Szecsenyi, and Matej Sandor for their help and valuable insights. The work of BS is supported by the UKRI EPSRC through the CDT in Machine Learning Systems hosted in the School of Informatics, University of Edinburgh (EP/Y03516X/1). The work of AS is supported by the UK Engineering and Physical Sciences Research Council (grant EP/W002876/1). ESW acknowledges support from the CIFAR Learning in Machines and Brains programme. This research was enabled by computer resources provided by the School of Informatics at the University of Edinburgh.

\bibliography{collas2026_conference}
\bibliographystyle{collas2026_conference}

\newpage
\appendix
\onecolumn
\section*{Appendix}

The appendix is organised as follows:
\Cref{app:notation} provides the notation used throughout the paper,
\Cref{app:theory} provides a formal theoretical exposition along with proofs,
\Cref{app:thought_exps} gives illustrative thought experiments,
\Cref{app:empirical} provides additional details regarding empirical implementation,
\Cref{app:results} reports additional results, and
\Cref{app:implementation} details our experiments.

\medskip

\begin{sc}
 \startcontents[appendix]
 \printcontents[appendix]{l}{1}{}
\end{sc}

\vfill
\begin{center}
    --appendices continue on next page--
\end{center}

\clearpage

\section{Notation}
\label{app:notation}
We first provide a table summarising the notation used throughout the paper.
\begin{table}[H]
\centering
\caption{A summary of notation.}
\begin{tabularx}{\linewidth}{lXp{5cm}}
\toprule
\textbf{Symbol} & \textbf{Description} & \textbf{Domain / Type} \\
\midrule
\multicolumn{3}{l}{\textbf{Interface-related}} \\
$(\mathcal X,\mathcal Y)$ & Environment–learner interface & Pair of measurable spaces \\
$\mathcal X$ & Observation space & Measurable space \\
$\mathcal Y$ & Prediction space & Measurable space \\
$\mathcal H$ & Set of histories & $\bigcup_{t\in\mathbb N_0} (\mathcal X\times \mathcal Y)^{t+1}$ \\
$X_t$ & Observation at time $t$ & Random variable \\
$Y_t$ & Output at time $t$ & Random variable \\
$H_{0:t}$ & History up to time $t$ & Sequence $((X_0,Y_0),\dots,(X_t,Y_t))$ \\
\midrule
\multicolumn{3}{l}{\textbf{Environment-related}} \\
$(e,p_{X_0})$ & Environment & $e:\mathcal H\to\mathcal{P}(\mathcal X),\;p_{X_0}\in\mathcal{P}(\mathcal X)$ \\
$p_e(\cdot\mid h,y)$ & Environment distribution given history $h$ and output $y$ & $\mathcal{P}(\mathcal X)$ \\
$q_e(\cdot\mid h,y)$ & Observation-generation kernel & $\mathcal{P}(\mathcal X)$ \\
\midrule
\multicolumn{3}{l}{\textbf{Learner-related}} \\
$(\mathcal Z,f,u,u',p_{Z_0})$ & Learner & Tuple \\
$\mathcal Z$ & Learner state space & Space \\
$Z_t$ & Learner state at time $t$ & Random variable \\
$Z_t^s$ & Learner state in inference mode at a futures time $s$ & Random variable \\
$p_{Z_0}$ & Initial learner state distribution & $\mathcal{P}(\mathcal Z)$ \\
$f$ & Prediction function & $\mathcal Z\times\mathcal X\to\mathcal{P}(\mathcal Y)$ \\
$u$ & Learning-mode state update & $\mathcal Z\times\mathcal X\times\mathcal Y\to\mathcal{P}(\mathcal Z)$ \\
$u'$ & Inference-mode state update & $\mathcal Z\times\mathcal X\times\mathcal Y\to\mathcal{P}(\mathcal Z)$ \\
$q_f(\cdot\mid z,x)$ & Predictive distribution & $\mathcal{P}(\mathcal Y)$ \\
\midrule
\multicolumn{3}{l}{\textbf{Predictive distributions}} \\
$q(H^{t+1:\infty}\mid Z_{t},H_{0:t})$ & Predictive distribution & $\mathcal{P}((\mathcal X\times\mathcal Y)^{\mathbb N})$ \\
$X^s$ & Future observation at introspection time $s$ & Random variable \\
$Y^s$ & Future output at introspection time $s$ & Random variable \\
$H^{t+1:\infty}$ & Sampled induced future at time $t$ & \parbox[t]{\hsize}{Sequence in\\$((X^{t+1},Y^{t+1}),(X^{t+2},Y^{t+2}),\dots)$} \\
\bottomrule
\end{tabularx}
\end{table}

\clearpage

\subsection{Interaction process generality}\label{app:framework_generality}
We now provide explicit examples of different realisations of the abstract variables introduced in \Cref{sec:prelims}.

\begin{table}[ht]
\centering
\caption{This table provides specific examples of values for the abstract variables: $X_t,Y_t$, and $Z_t$ across different common machine learning settings.}
\begin{tabularx}{\linewidth}{>{\raggedright\arraybackslash}p{2.2cm}YYY}
\toprule
\textbf{Setting} & \textbf{$X_t$ (observation)} & \textbf{$Y_t$ (prediction)} & \textbf{$Z_t$ (learner state)} \\
\midrule
\makecell[l]{\\Regression} & Current input, previous target pair $(x_t,\;y_{t-1})$ & Predicted target $\hat{y}_{t-1}$ & Parameters (e.g., weights, optimiser state) \\
\makecell[l]{\\Classification} & Current input, previous label pair $(x_t,\;\ell_{t-1})$ & Predicted class label $\hat{\ell}_{t-1}$ & Classifier parameters, latent state (e.g., neural network weights, momenta) \\
\makecell[l]{\\Generative\\modelling} & New observation, previous target generation $(x_t,\;y_{t-1})$ & Attempted generations of $x_{t-1}$ & Generative model parameters, latent variables, buffers \\
\makecell[l]{\\Bayesian\\models} & Data (possibly $(x_t,y_{t-1})$ pairs) & Posterior predictive sample $\hat y_{t-1}$ & Posterior / variational parameters, latent variables\newline \\
\makecell[l]{Reinforcement\\learning} & \makecell[l]{State–reward $(s,r)$ \\ or transition $(s,a,r,s')$} & \makecell[l]{Action $a$ (policy output) \\ or predictive model of \\ next state} & \makecell[l]{Policy/value parameters, \\ transition/reward models, \\ replay buffer, auxiliary \\ structures} \\
\bottomrule
\end{tabularx}
\label{tab:framework}
\end{table}

\vfill
\begin{center}
    --appendices continue on next page--
\end{center}

\clearpage

\section{Theoretical results}\label{app:theory}https://www.overleaf.com/project
Recall, the learner's state is updated according to
\begin{equation}
    Z_t\sim u(\cdot\mid Z_{t-1},X_t,Y_t).
\end{equation}
Before observing $X_t,Y_t$ and updating to $Z_t$, we have a futures distribution
\begin{equation}
    q(H^{t+1:\infty}\mid Z_{t-1},H_{0:t-1}).
\end{equation}

\subsection{One-step simulated marginalisation}\label{app:one_step}
The one-step simulated marginalisation is obtained by marginalising over observations $X_t,Y_t$, and the updated state $Z_t$:
\begin{equation}
    q(H^{t+1:\infty}\mid Z_{t-1},H_{0:t-1})=\int q(H^{t+1:\infty},X_t,Y_t,Z_t\mid Z_{t-1},H_{0:t-1})\,\mathrm d X_t\,\mathrm dY_t\,\mathrm dZ_t.
\end{equation}
This can be factorised to:
\begin{equation}
    \begin{aligned}
        &\int q(X_t,Y_t,Z_t,H^{t+1:\infty}\mid Z_{t-1},H_{0:t-1})\,\mathrm d X_t\,\mathrm dY_t\,\mathrm dZ_t\\
        &=\int q(H^{t+1:\infty}\mid X_t,Y_t,Z_{t},Z_{t-1},H_{0:t-1})\,u(Z_t\mid Z_{t-1},X_t,Y_t,H_{0:t-1})\\
        & \qquad\qquad\cdot q_e(X_t\mid Y_t,Z_{t-1},H_{0:t-1})\,q_f(Y_t\mid Z_{t-1},H_{0:t-1})\,\mathrm d X_t\,\mathrm dY_t\,\mathrm dZ_t.
    \end{aligned}
\end{equation}
In generic learning scenarios and under the assumption of learning over time, we assume no access to the history.
Thus, the simulated marginalisation ensures that the state update, input, and prediction are conditionally independent of the history.
Furthermore, the induced future is conditionally independent of past learner states:
\begin{equation}
  \begin{aligned}
    X_t&\indep H_{0:t-1}\mid Y_t,Z_{t-1},\qquad\quad\,\, Y_t\indep (H_{0:t-2}, Y_{t-1})\mid Z_{t-1},X_{t-1}\\
    Z_t&\indep H_{0:t-1}\mid X_t,Y_t,Z_{t-1},\qquad H^{t+1:\infty}\indep Z_{t-1}\mid Z_t.
  \end{aligned}
\end{equation}
This reduces the factorised form to:
\begin{equation}\label{eq:one_step_expanded} 
    \begin{aligned}
        &=\int q(H^{t+1:\infty}\mid Z_{t},H_{0:t})\,u(Z_t\mid Z_{t-1},X_t,Y_t)\\
        & \qquad\qquad\cdot q_e(X_t\mid Y_t,Z_{t-1})\,q_f(Y_t\mid Z_{t-1},X_{t-1})\,\mathrm d X_t\,\mathrm dY_t\,\mathrm dZ_t.
    \end{aligned}
\end{equation}
which is the 1-step simulated marginalisation:
\begin{equation}\label{eq:one_step_consistency_appendix}
    q^*_1(H^{t+1:\infty}\mid Z_{t-1},H_{0:t-1})=\mathbb{E}_{X_t,Y_t,Z_t}\left[ q(H^{t+1:\infty}\mid Z_t,H_{0:t})\right],
\end{equation}
where
\begin{equation*}
  X_t\sim q_e(\cdot\mid Y_t,Z_{t-1}), \quad Y_t\sim q_f(\cdot\mid Z_{t-1},X_{t-1}), \quad Z_t\sim u(\cdot\mid Z_{t-1},X_t,Y_t).
\end{equation*}

\subsection{k-step simulated marginalisation}\label{app:k_step}
We can apply the same reasoning to derive the $k$-step simulated marginalisation.
If we denote $t'$ as $t'=t+k-1$, we get:
\begin{equation}
    \begin{aligned}
        &q(H^{t+k:\infty}\mid Z_{t-1},H_{0:t-1})\\
        &=\int q(X_{t:t'},Y_{t:t'},Z_{t:t'},H^{t+k:\infty}\mid Z_{t-1},H_{0:t-1})\,\mathrm d X_{t:t'}\,\mathrm dY_{t:t'}\,\mathrm dZ_{t:t'}\\
        &=\int q(H^{t+k:\infty}\mid Z_{t'},H_{0:t'})\cdot\prod_{s=t}^{t'} \big[u(Z_s\mid Z_{s-1},X_s,Y_s)\\
        &\qquad\cdot q_e(X_s\mid Y_s, Z_{s-1})q_f(Y_s\mid Z_{s-1},X_{s-1})\big]\mathrm d X_{t:t'}\,\mathrm dY_{t:t'}\,\mathrm dZ_{t:t'},
    \end{aligned}
\end{equation}
which is the $k$-step simulated marginalisation given in \Cref{def:consistency_condition}:
\begin{equation}
    q_k^*(H^{t+k:\infty}\mid Z_{t-1},H_{0:t-1})=\mathbb{E}_{X_{t:t'},Y_{t:t'},Z_{t:t'}}\left[ q(H^{t+k:\infty}\mid Z_{t'},H_{0:t'})\right].
\end{equation}
where, for $i=t,\dots,t'$
\begin{equation*}
  X_i\sim q_e(\cdot\mid Y_i, Z_{i-1}), \quad Y_i\sim q_f(\cdot\mid Z_{i-1},X_{i-1}), \quad Z_i\sim u(\cdot\mid Z_{i-1},X_i,Y_i).
\end{equation*}

\subsection{Bayesian non-forgetting}\label{app:bayesian_theorem}
We now formally state and prove \Cref{thm:bayes_non_forget}.

Let $\Theta$ denote a measurable latent space that takes values $\theta\in\Theta$.
The learner maintains a state $Z_t$ for $t\ge0$, which defines a posterior distribution $q_{Z_t}$ over $\Theta$. We assume the existence of predictive distributions $q(H^{t+1:\infty}\mid \theta,X_t,Y_t)$ for $\theta\in\Theta$, which are related to the predictive distributions conditioned on $Z_t$ (\Cref{sec:futures}) via \Cref{assumption:mixture} below.

In what follows, we assume that the prior and all posterior measures are absolutely continuous with respect to some fixed $\sigma$-finite base measure $\mu$ on $\Theta$. This assumption justifies the use of distributions over $\Theta$ interchangeably with their densities (Radon-Nikodym derivatives with respect to $\mu$), which are unique up to $\mu$-a.e.\ equality.

We further remind the reader of the contents of $X_t,Y_t$ and $Z_t$.
$X_t$ is the observation made by the learner at time $t$, $Y_t$ is the output of the learner on the previous observation $Y_t\mid X_{t-1}$, and $Z_t$ is the learner's state, including things like parameters, memory buffers, and other auxiliary components.
Thus, in a supervised setting, $X_t$ contains both the current input and the previous target, $X_t=(x_t,y_{t-1})$.


We will show that under simple assumptions, the posterior predictive is \emph{self-consistent} under Bayesian updates.
This implies that, under Bayesian updates, the simulated marginalisation yields the same predictive distribution as the original predictive distribution.
\begin{assumption}[Likelihood Existence]\label{assumption:likelihood}
    For any $t$, the marginal likelihood obtained by marginalising over $\theta$,
    \begin{equation}
      q_e(X_t\mid Y_t,Z_{t-1})=\int_\Theta q(X_t\mid Y_t,\theta)q_{Z_{t-1}}(\theta)\,\mathrm d\theta,
    \end{equation}
    is well defined and finite, with the integrand integrable over $\Theta$ with respect to $q_{Z_{t-1}}(\theta)\,\mathrm d\theta$.
\end{assumption}
\begin{assumption}[Bayesian Updates]\label{assumption:updates}
    The learner's posterior, denoted by $q_{Z_t}(\theta)\coloneqq q(\theta\mid Z_t)$, is updated by Bayes' rule
    \begin{equation}\label{eq:bayes_update}
      \begin{aligned}
        q_{Z_t}(\theta)=\frac{q(X_t\mid Y_t,\theta)q_{Z_{t-1}}(\theta)}{q_e(X_t\mid Y_t, Z_{t-1})},\\
      \end{aligned}
    \end{equation}
    where $q_e(X_t\mid Y_t, Z_{t-1})$ is given by \Cref{assumption:likelihood}.
    Therefore, the update procedure is deterministic.
    We denote the $Z_t$ that represents the resulting distribution by $\Phi(Z_{t-1},X_t,Y_t)$, so the update rule is
    \begin{equation}
        u(\cdot\mid Z_{t-1},X_t,Y_t)=\delta_{\Phi(Z_{t-1},X_t,Y_t)}.
    \end{equation}
\end{assumption}
\begin{assumption}[Predictive Conditional Independence]\label{assumption:markov}
    During simulated marginalisation, the predictive distribution $H^{t+1:\infty}$ is generated by the current state $Z_t$, and is therefore conditionally independent of past states $Z_0,Z_t,\dots,Z_{t-1}$:
    \begin{equation}
        q(H^{t+1:\infty}\mid X_t,Y_t,Z_{t},Z_{t-1},H_{0:t-1})=q(H^{t+1:\infty}\mid X_t,Y_t,Z_{t},H_{0:t-1}).
    \end{equation}
\end{assumption}
\begin{assumption}[Mixture Representation]\label{assumption:mixture}
    For any $t$, the learner's predictive future distribution is equal to the mixture
    \begin{equation}
      q(H^{t+1:\infty}\mid Z_t, X_t, Y_t)=\int_\Theta q(H^{t+1:\infty}\mid \theta, X_t, Y_t)q_{Z_t}(\theta)\,\mathrm d\theta.
    \end{equation}
\end{assumption}
\begin{assumption}[Regularity]\label{assumption:regularity}
    All probability distributions are non-negative measurable functions and may be integrated over the considered ranges.
    Thus, Tonelli's/Fubini's theorem \citep{tao2006analysis} may be applied to change the order of integration.
\end{assumption}

We first show that exact Bayesian updates do not forget after an update.
\begin{lemma}[One-Step Bayesian Non-Forgetting]\label{thm:bayes_non_forget_1step}
  Under \Cref{assumption:updates,assumption:markov,assumption:mixture,assumption:likelihood,assumption:regularity}, for every $t\in\mathbb N$, the learner's posterior-predictive distribution satisfies the one-step self-consistency condition,
    \begin{equation}
      q(H^{t+1:\infty}\mid Z_{t-1},H_{0:t-1})=q^*_1(H^{t+1:\infty}\mid Z_{t-1},H_{0:t-1}),
    \end{equation}
    where state updates are determined by deterministic Bayesian updates.
    Thus, $\Gamma_1(t)=0$ for all $t\in\mathbb N$.
\end{lemma}
\begin{proof}
  Start from the one-step simulated marginalisation expansion in \eqref{eq:one_step_expanded}:
    \begin{equation}
      \begin{aligned}
        q^*_1(H^{t+1:\infty}\mid Z_{t-1},H_{0:t-1})
                                               =\iiint &q(H^{t+1:\infty}\mid Z_{t},H_{0:t})\\
                                               &\cdot u(Z_t\mid Z_{t-1},X_t,Y_t)\\
                                               &\cdot q_e(X_t\mid Y_t,Z_{t-1})\\ 
                                               &\cdot q_f(Y_t\mid Z_{t-1}, X_{t-1})\,\mathrm d X_t\,\mathrm dY_t\,\mathrm dZ_t.
      \end{aligned}
    \end{equation}
    Under \Cref{assumption:updates}, the update is deterministic: $u(Z_t\mid Z_{t-1},X_t,Y_t)=\delta_{\Phi(\cdot)}$.
    Therefore, the triple integral reduces to a double integral:
    \begin{equation}
      \begin{aligned}
        q^*_1(H^{t+1:\infty}\mid Z_{t-1},H_{0:t})&=\iint q(H^{t+1:\infty}\mid Z_t=\Phi(Z_{t-1},X_t,Y_t),H_{0:t})\\
        & \qquad\quad\cdot q_e(X_t\mid Y_t,Z_{t-1})q_f(Y_t\mid Z_{t-1}, X_{t-1})\,\mathrm d X_t\,\mathrm dY_t.
      \end{aligned}
    \end{equation}
    Substituting \Cref{assumption:mixture} into the double integral and using \Cref{assumption:regularity} to swap the order of integration gives
    \begin{equation}
      \begin{aligned}
        &=\int\left[\iint q(H^{t+1:\infty}\mid\theta,H_{0:t})q_{Z_t}(\theta)\right. \\
        & \qquad\qquad\cdot q_e(X_t\mid Y_t,Z_{t-1})q_f(Y_t\mid Z_{t-1}, X_{t-1})\,\mathrm d X_t\,\mathrm dY_t\bigg]\,\mathrm d\theta.
      \end{aligned}
    \end{equation}
    Inserting the Bayes update formula (\Cref{assumption:updates}) for $q_{Z_t}(\theta)$ gives:
    \begin{align}\label{eq:30}
        &=\int\left[\iint q(H^{t+1:\infty}\mid\theta,H_{0:t})\frac{q(X_t\mid Y_t,\theta)q_{Z_{t-1}}(\theta)}{q_e(X_t\mid Y_t, Z_{t-1})}\right. \\
        & \qquad\qquad\cdot q_e(X_t\mid Y_t,Z_{t-1})q_f(Y_t\mid Z_{t-1}, X_{t-1})\,\mathrm d X_t\,\mathrm dY_t\bigg]\,\mathrm d\theta\nonumber\\
        &=\int\left[\iint q(H^{t+1:\infty}\mid\theta,H_{0:t})q_{Z_{t-1}}(\theta)
        \underbrace{q(X_t\mid Y_t,\theta)
        q_f(Y_t\mid Z_{t-1}, X_{t-1})}_{=q(X_t, Y_t \mid Z_{t-1}, X_{t-1},\theta)}
        \,\mathrm d X_t\,\mathrm dY_t
        \right]\,\mathrm d\theta,
    \end{align}
    where the joint distribution is derived from the conditional independence properties discussed in \Cref{app:one_step}.
    By applying the chain rule of probability, changing the order of integration (\Cref{assumption:regularity}), and then using the equality in \Cref{assumption:mixture}, we get
    \begin{equation}
      \begin{aligned}
        &=\iint\left[\int q(H^{t+1:\infty},X_t,Y_t\mid\theta,Z_{t-1},H_{0:t-1})q_{Z_{t-1}}(\theta)\,\mathrm d\theta \right] \,\mathrm d X_t\,\mathrm dY_t\\
        &=\iint q(H^{t+1:\infty},X_t,Y_t\mid Z_{t-1},H_{0:t-1})\,\mathrm d X_t\,\mathrm dY_t\\
        &=q(H^{t+1:\infty}\mid Z_{t-1},H_{0:t-1}).
      \end{aligned}
    \end{equation}
    Thus, even when the learner has no access to the history, the Bayesian inference procedure does not forget under the above assumptions:
    \begin{equation}\label{eq:equal}
      q(H^{t+1:\infty}\mid Z_{t-1},H_{0:t-1})=q^*_1(H^{t+1:\infty}\mid Z_{t-1},H_{0:t-1}),
    \end{equation}
    completing the proof.

    Given the equality in \eqref{eq:equal}, for every choice of divergence measure, $\Gamma_1(t)=0$.
\end{proof}

Now that we have shown the one-step non-forgetting property of Bayesian updates (\Cref{thm:bayes_non_forget_1step}), we can extend the result to the general $k$-step self-consistency condition.
First, we restate \Cref{thm:bayes_non_forget}.
\begin{theoremnum}[Bayesian Non-Forgetting]{thm:ftc}{5.1}
  Let $k\in\mathbb N$ and for every $t\in\mathbb N$ define $t'=t+k-1$. Under \Cref{assumption:updates,assumption:markov,assumption:mixture,assumption:likelihood,assumption:regularity}, the learner's posterior-predictive distribution satisfies
    \begin{equation}
      q(H^{t+k:\infty}\mid Z_{t-1},H_{0:t-1})=q_{k}^*(H^{t+k:\infty}\mid Z_{t-1},H_{0:t-1})
    \end{equation}
    when $Z_t$ is determined by deterministic Bayesian updates.
    Thus, $\Gamma_k(t)=0$ for all $k,t\in\mathbb N$.
\end{theoremnum}

\begin{proof}
  We prove by induction.
  The base case ($k=1$) is \Cref{thm:bayes_non_forget_1step}:
  \begin{equation}
      q(H^{t+1:\infty}\mid Z_{t-1},H_{0:t-1})=q^*_1(H^{t+1:\infty}\mid Z_{t-1},H_{0:t-1}),
  \end{equation}
  We now prove the inductive step.

  Let $t'=t+k-1$.
  Assume the $k$-step self-consistency condition holds:
  \begin{equation}\label{eq:44}
    \begin{aligned}
      q(H^{t+k:\infty}\mid Z_{t-1},H_{0:t-1})=q^*_k(H^{t+k:\infty}\mid Z_{t-1},H_{0:t-1}).
    \end{aligned}
  \end{equation}
  We wish to show that \eqref{eq:44} holds for $k+1$.

  We begin with the $k+1$-step self-consistency expansion:
  \begin{equation}
    \begin{aligned}
      q_{k+1}^*(H^{t+k+1:\infty}\mid Z_{t-1},&H_{0:t-1})\\
      =\iiint &q(H^{t+k+1:\infty}\mid Z_{t'+1},H_{0:t'+1})\\
              &\cdot \prod_{s=t}^{t'+1} \bigg[u(Z_s\mid Z_{s-1},X_s,Y_s)q_e(X_s\mid Y_s,Z_{s-1})\\
               &\cdot q_f(Y_s\mid Z_{s-1},X_{s-1})\bigg]\,\mathrm d X_{t:t'+1}\,\mathrm dY_{t:t'+1}\,\mathrm dZ_{t:t'+1}.
    \end{aligned}
  \end{equation}
  The $t'+1$ updates are deterministic $u(Z_{t'+1}\mid Z_{t'},X_{t'+1},Y_{t'+1})=\delta_{\Phi(\cdot)}$ under \Cref{assumption:updates}; therefore, they can be represented as an iterative deterministic update from $Z_{t-1}$:
  \begin{equation}
    \begin{aligned}
      =\iint &q(H^{t+k+1:\infty}\mid Z_{t'+1} =\Phi(\Phi(\dots\Phi(Z_{t-1},X_t,Y_t)\dots),X_{t'+1},Y_{t'+1}),H_{0:t'+1})\\
              &\cdot\prod_{s=t}^{t'+1} \bigg[q_e(X_s\mid Y_s,Z_{s-1})q_f(Y_s\mid Z_{s-1},X_{s-1})\bigg]\,\mathrm d X_{t:t'+1}\,\mathrm dY_{t:t'+1}.
    \end{aligned}
  \end{equation}
  Substituting \Cref{assumption:mixture} for the predictive distribution and using \Cref{assumption:regularity} to swap the order of integration gives
  \begin{equation}
    \begin{aligned}
      =\iiint &q(H^{t+k+1:\infty}\mid \theta,H_{0:t'+1})q_{Z_{t'+1}}(\theta)\prod_{s=t}^{t'+1} \bigg[q_e(X_s\mid Y_s,Z_{s-1})q_f(Y_s\mid Z_{s-1},X_{s-1})\bigg]\,\mathrm d X_{t:t'+1}\,\mathrm dY_{t:t'+1}\,\mathrm d\theta.
    \end{aligned}
  \end{equation}
  Inserting the Bayes update formula (\Cref{assumption:updates}) for $q_{Z_{t'+1}}(\theta)$ gives:
  \begin{equation}\label{eq:41}
    \begin{aligned}
      =\iiint &q(H^{t+k+1:\infty}\mid \theta,H_{0:t'+1})\frac{q_{Z_{t-1}}(\theta)\prod_{s=t}^{t'+1}q(X_s\mid Y_s,\theta)}{\prod_{s=t}^{t'+1}q_e(X_s\mid Y_s,Z_{s-1})}\\
                        &\cdot \prod_{s=t}^{t'+1} \bigg[q_e(X_s\mid Y_s,Z_{s-1})q_f(Y_s\mid Z_{s-1},X_{s-1})\bigg]\,\mathrm d X_{t:t'+1}\,\mathrm dY_{t:t'+1}\,\mathrm d\theta\\
      =\iiint &q(H^{t+k+1:\infty}\mid \theta,H_{0:t'+1})q_{Z_{t-1}}(\theta)\prod_{s=t}^{t'+1}\bigg[q(X_s\mid Y_s,\theta)q_f(Y_s\mid Z_{s-1},X_{s-1})\bigg]\,\mathrm d X_{t:t'+1}\,\mathrm dY_{t:t'+1}\,\mathrm d\theta,\\
    \end{aligned}
  \end{equation}
  where $Z_{s-1}$ is deterministically obtained by recursively applying the Bayesian update $\Phi$ to the given state $Z_{t-1}$ and the marginalised data $(X_t,Y_t),\dots,(X_{t'+1},Y_{t'+1})$.
  We denote this progression as $Z_{s-1}=\Phi_s(Z_{t-1},X_{t:s-1},Y_{t:s-1})$, with base case $\Phi_t(Z_{t-1})=Z_{t-1}$:
  \begin{equation}
    \begin{aligned}
      =\iiint &q(H^{t+k+1:\infty}\mid \theta,H_{0:t'+1})q_{Z_{t-1}}(\theta)\\
                       &\cdot\underbrace{q(X_t\mid Y_t,\theta)q_f(Y_t\mid Z_{t-1},X_{t-1})\prod_{s=t+1}^{t'+1}\bigg[q(X_s\mid Y_s,\theta)q_f(Y_s\mid \Phi_s(Z_{t-1},X_{t:s-1},Y_{t:s-1}),X_{s-1})\bigg]}_{q(X_{t:t'+1},Y_{t:t'+1}\mid Z_{t-1},X_{t-1},\theta)}\\
                       &\mathrm d X_{t:t'+1}\,\mathrm dY_{t:t'+1}\,\mathrm d\theta.\\
    \end{aligned}
  \end{equation}
  By applying the chain rule of probability, we get
  \begin{equation}
    \begin{aligned}
      =\iiint &q(H^{t+k+1:\infty}\mid \theta,H_{0:t'+1})q_{Z_{t-1}}(\theta)q(X_{t:t'+1},Y_{t:t'+1}\mid Z_{t-1},X_{t-1},\theta)\,\mathrm d X_{t:t'+1}\,\mathrm dY_{t:t'+1}\,\mathrm d\theta\\
      =\iiint &q(H^{t+k+1:\infty},X_{t:t'+1},Y_{t:t'+1}\mid \theta,Z_{t-1},H_{0:t-1})q_{Z_{t-1}}(\theta)\,\mathrm d X_{t:t'+1}\,\mathrm dY_{t:t'+1}\,\mathrm d\theta.\\
    \end{aligned}
  \end{equation}
  Using \Cref{assumption:regularity} to swap the order of integration gives:
  \begin{equation}
    \begin{aligned}
      &=\iiint q(H^{t+k+1:\infty},X_{t:t'+1},Y_{t:t'+1}\mid \theta,Z_{t-1},H_{0:t-1})q_{Z_{t-1}}(\theta)\,\mathrm d\theta\,\mathrm d X_{t:t'+1}\,\mathrm dY_{t:t'+1}\\
      &=\iint q(H^{t+k+1:\infty},X_{t:t'+1},Y_{t:t'+1}\mid Z_{t-1},H_{0:t-1})\,\mathrm d X_{t:t'+1}\,\mathrm dY_{t:t'+1}.\\
      &=q(H^{t+k+1:\infty}\mid Z_{t-1},H_{0:t-1}).
    \end{aligned}
  \end{equation}
  Thus, the equality
  \begin{equation}\label{eq:50}
    \begin{aligned}
      q(H^{t+k+1:\infty}\mid Z_{t-1},H_{0:t-1})=q_{k+1}^*(H^{t+k+1:\infty}\mid Z_{t-1},&H_{0:t-1})
    \end{aligned}
  \end{equation}
  holds for $k+1$.
  By induction, this holds for all $k\in\mathbb N$.

  Given the equality in \eqref{eq:50}, for every choice of divergence measure, $\Gamma_k(t)=0$.
\end{proof}

\paragraph{Implications.}
The non-forgetting result relies on a set of assumptions about the learner's inference procedure and interaction dynamics.
\Cref{assumption:updates} requires that the posterior $q_{Z_t}$ is obtained from a deterministic Bayesian update, ensuring the posterior $q_{Z_t}$ is an exact update of the preceding posterior $q_{Z_{t-1}}$. 

This assumption is not satisfied by approximate inference procedures, such as variational Bayes, Laplace approximations, and related approaches. In these approximate inference procedures, the self-consistency condition is not guaranteed to hold, implying that such approaches exhibit forgetting.

\subsection{Martingale perspective}\label{app:martingale}
Information available over time may be formalised by a filtration $(\mathcal F_t)_{t\ge0}$, where $\mathcal F_t$ is a $\sigma$-algebra representing all information observable up to time $t$.
In this construction, once an observation is made, the information it contains is never lost \citep{doob1953stochastic}.

In our setting, let
\begin{equation}
  \mathcal F_t\coloneqq \sigma(X_{0:t},Y_{0:t},Z_{0:t})
\end{equation}
denote the full interaction filtration generated by the history of observations, outputs, and learner states up to time $t$.

The learner does not model the entire interaction process.
Instead, all information retained by the learner is summarised by its state $Z\in\mathcal Z$.
Therefore, we consider the \emph{state filtration}
\begin{equation}
  \mathcal G_t\coloneqq \sigma(Z_0,Z_1,\dots,Z_t),
\end{equation}
which encodes all of the information retained by the learner up to time $t$.

To define a martingale at a fixed interaction step $t$, we consider a fixed future horizon $k\in\mathbb{N}$ and define the predictive distribution
\begin{equation}
  Q^0_k\coloneq q(H^{t+k:\infty}\mid Z_{t},H_{0:t}).
\end{equation}

To analyse forgetting, we study how this predictive distribution evolves under simulated marginalisation.
Let $(Z^s_t)_{s\in[0,k]}$ denote the sequence of learner states obtained by applying the learner's transition dynamics for $s\in\mathbb N_0$ simulated steps, starting from $Z^0_t=Z_t$.
The associated filtration over simulated time is
\begin{equation}
  \mathcal G^s\coloneqq \sigma(Z^0_t,\dots,Z^s_t).
\end{equation}
For each simulated step $s$, define the corresponding predictive random variable as
\begin{equation}
  Q^s_k\coloneqq q(H^{t+k:\infty}\mid Z^s_t,H_{0:t}).
\end{equation}
The learner does not forget over $k$ steps if and only if it holds that 
\begin{equation}
  \mathbb E[Q^k_k\mid\mathcal G^0]=Q^0_k.
\end{equation}
A sufficient condition for non-forgetting is that the process $(Q^s_k)_{s\in[0,k]}$ forms a martingale with respect to the simulated filtration $(\mathcal G^s)_{s\in[0,k]}$:
\begin{equation}\label{eq:martingale}
  \mathbb E[Q^{s+1}_k\mid\mathcal G^s]=Q^s_k,\qquad \text{for all }s\in[0,k-1].
\end{equation}
If \eqref{eq:martingale} does not hold, then the learner's predictive beliefs drift during the simulated marginalisation.
This indicates that the learner has lost information contained within its previous states, i.e., the learner has forgotten.
The expectation in \eqref{eq:martingale} is taken with respect to the learner's simulated marginalisation, isolating forgetting from backward transfer.

Approximate learners typically violate \eqref{eq:martingale}, thereby losing predictive information.
As shown in \Cref{app:bayesian_theorem}, exact Bayesian updates preserve self-consistency and therefore satisfy the martingale property.
Therefore, they do not forget.

\subsection{The need for replay}\label{app:replay-theory}
The above derivations make the role of the learner's state update function, $u(\cdot\mid Z_{t-1}, X_t, Y_t)$, during the simulated marginalisation explicit. The learner's update must be such that $Z_t$ is conditionally independent of the whole history $H_{0:t-1}$ given $(Z_{t-1},X_t,Y_t)$, and such that the predictive futures $H^{t+k:\infty}$ are conditionally independent of all of the previous states during the updates $Z_t,Z_{t+1},\dots,Z_{t+k-2}$ given the current state $Z_{t+k-1}$. 

However, in practice, many learning algorithms violate these conditions. When these dependencies exist, performing consistent updates during simulated marginalisation depends on access to past data or states. Replay mechanisms provide an empirical solution: by placing past observations into the current observation, the learner effectively reduces its dependence on previous states and the history.

\vfill
\begin{center}
    --appendices continue on next page--
\end{center}

\clearpage

\section{Thought experiments}\label{app:thought_exps}
We present a series of thought experiments designed to stress-test potential definitions of forgetting in general learning systems. Each scenario illustrates a conceptual edge case that any robust definition of forgetting should naturally handle. For each scenario, we provide a \emph{self-consistency verdict}, indicating whether our framework classifies the learner as forgetting or not, and we explain the intuition behind this judgement.

\begin{scenario}[The degenerate learner.]
Consider a learner that never updates its beliefs — its state is fixed for all time. Such a learner never changes its state or acquires new information and is thus degenerate. Does this learner ever forget? One might argue that forgetting is defined relative to the information available to the learner over time; however, if nothing is ever learned, how can anything ever be forgotten?

\emph{Self-consistency verdict: no forgetting.} Since the state remains constant across any number of updates, the predictive distribution is identical to the simulated marginalisation. A learner that never changes its state can never forget.
\end{scenario}

\begin{scenario}[0-bit and 5-bit stacks.]
Consider two first-in, first-out (FIFO) stacks: one with capacity 0 bits, another with 5 bits. The 0-bit stack cannot store information; the 5-bit stack can.

\emph{Self-Consistency verdict: 0-bit does not forget, the 5-bit stack does.} The predictive distribution of the stack is always the returned bit of the address being accessed at each step. A 0-bit stack's state never changes, so its future remains constant. The 5-bit stack, however, overwrites past bits when it is full; its predictive distribution changes as bits are dropped, hence it forgets. Forgetting only exists in systems capable of learning.
\end{scenario}

\begin{scenario}[The hash map.]
A hash map with infinite associative memory never overwrites existing entries. It represents a theoretical ``perfect rememberer".

\emph{Self-Consistency verdict: no forgetting.} The predictive distributions of such a learner are a Dirac at the retrieved value (or null if unassigned), and the predictive distribution and simulated marginalisation remain identical over time. Thus, the learner never forgets.
\end{scenario}

\begin{scenario}[The clock.]
Consider a simple clock that increments its state deterministically with each tick. Does the clock forget past states? In one sense, yes: it overwrites its register and never recovers earlier times. In another sense, no: it never misrepresents the current time and updates consistently. This raises the question: must forgetting be tied to the loss of recoverable information, or only to deviations from self-consistent updates?

\emph{Self-Consistency verdict: no forgetting.} Under learning environmental updates, the clock advances its state (its notion of ``time") by one unit per tick. For any $k$, the time from $t+k:\infty$ under simulated updates matches that under learning interaction. Consequently, the learner's predicted futures remain invariant under self-consistent updates. The clock does not forget.
\end{scenario}

\begin{scenario}[The moody learner.]
Suppose a learner updates only on even time steps, ignoring all odd ones. Does inaction indicate perfect memory?

\emph{Self-Consistency verdict: no forgetting (on ignored steps).} When no update occurs, the state remains identical, so the predictive distribution is equivalent to the simulated marginalisation. Forgetting can only occur during state-altering updates.
\end{scenario}

\begin{scenario}[The function picker.]
Suppose there are $L$ learner functions, and at each timestep, one function is selected uniformly at random, independent of both $t$ and the observed data.

\emph{Self-Consistency verdict: no forgetting.} In expectation, the predictive distribution after any number of self-consistent updates remains invariant. The randomness of selection does not imply loss of information about possible futures.
\end{scenario}

\begin{scenario}[The binary flipper.]
Consider a binary classification model that flips its predictions at every timestep.

\emph{Self-Consistency verdict: no forgetting.} Under self-consistent updates, all transitions up to time $t+k$ match those produced under learning interactions. Consequently, the binary flipper's futures remain invariant under self-consistent updates, and the flipper does not forget.
\end{scenario}

\begin{scenario}[Label permutation.]
Take a perfectly trained binary classifier. Now, consider permuting the mapping between logits and output labels without changing the parameters. On the one hand, the parameters remain the same; on the other, the outputs have changed significantly. Has such a model forgotten everything it knew just because its behaviour has changed, or has it not forgotten anything at all because it has the same states?

\emph{Self-Consistency verdict: this forgets.} During the simulated marginalisation, when a label permutation occurs, the predictive distribution changes, implying the learner has forgotten when a permutation occurs.
\end{scenario}

\begin{scenario}[Forgetting unseen but generalised inputs.]
Consider a learner that is trained on the MNIST dataset. At time $t$, the model correctly classifies all instances of the digit “4” in the test set. After further training, it loses the ability to correctly classify some test examples. Can a model forget something that it has never directly encountered (such as specific test inputs) but has nonetheless learned to generalise to?

\emph{Self-Consistency verdict: this forgets.} Even though the test data were unseen, the learner's predictive distribution changes when its performance on those data changes. Thus, the learner has forgotten: forgetting occurs whenever the previously supported predictive capabilities vanish, regardless of whether the corresponding data have been observed.
\end{scenario}

\begin{scenario}[Even number checkers.]
Suppose a model receives binary inputs sequentially and deterministically predicts a $1$ if it has observed an even number of $1$s, and $0$ otherwise. Does it forget because its state changes over time?

\emph{Self-Consistency verdict: no forgetting.} The next $k$ observations may either be a $0$ or a $1$, drawn from any distribution. After $k$ updates, the learner's predictive distribution is deterministically either a $0$ or a $1$. Although the state continues to evolve during the simulated marginalisation, the expected predictive distribution remains the same across all states. Therefore, no forgetting occurs.
\end{scenario}

\begin{scenario}[Surprising events.]\label{par:surprising}
Consider the scenario where a fair coin is flipped repeatedly. After 10 coin flips, we are in an unlikely situation where all flips have so far resulted in heads. If, on the next flip, we observe a tail, the learner may be surprised, even if it has remembered all previous results.

\emph{Self-Consistency verdict: no forgetting.} Under consistent updates, future updates are performed relative to the learner's current beliefs rather than the actual environment. This distinction is what prevents surprise from being mistaken for forgetting. Since the learner's predictive distribution already encodes its uncertainty about future outcomes, an unexpected tail does not constitute a loss of information; it is a valid observation under the learner's model. Consequently, the expected futures distribution before and after the $k$ updates remains identical, and no forgetting is detected.
\end{scenario}

\begin{scenario}[Bayesian optimisation.]
Consider a Bayesian optimiser that initially observes high objective values, then later finds a lower minimum. Its predictive distribution must update in response to the new observation. However, no previously acquired evidence has been discarded. This is a knowledge \emph{update}, not forgetting; naive comparisons of states may mistakenly classify this update as forgetting.

\emph{Self-Consistency verdict: no forgetting.} As in \Cref{par:surprising}, future updates are computed relative to the learner's current predictive distribution rather than the actual environment. Because updates reflect the learner's own beliefs, encountering a new minimum does not imply a loss of prior information. In expectation, the futures distribution before and after $k$ updates is identical, indicating that the learner has not forgotten.
\end{scenario}

\vfill
\begin{center}
    --appendices continue on next page--
\end{center}

\clearpage

\section{Empirical considerations}\label{app:empirical}
In this section, we detail the approach used to estimate the forgetfulness measure $\Gamma_k(t)$. $\Gamma_k(t)$ captures the extent to which the learner's induced predictive distribution at time $t$ remains consistent with those obtained after $k$ simulated updates. While the simulated marginalisation $q_k^*(\ldots)$ cannot be computed exactly, it can be approximated using a particle-based Monte Carlo scheme. At a given time, we clone the learner into $M$ replicas and propagate each replica forward over $k$ steps of simulated interaction, with each replica performing its own simulated marginalisation. This yields a set of post-update models $\{Z_{t+k-1}^{(m)}\}_{m=1}^M$ that allow us to approximate the predictive distribution with the following empirical distribution:
\begin{equation}
    q_k^*(H^{t+k:\infty}\mid Z_{t-1},H_{0:t-1})\approx \frac 1M \sum_{m=1}^Mq(H^{t+k:\infty}\mid Z_{t+k-1}^{(m)},H^{(m)}_{t+k-1}),
\end{equation}
We compare the learner's empirical predictive distribution after a simulated marginalisation with this empirical distribution. After selecting an appropriate divergence measure, we measure the divergence between the two empirical distributions to acquire an empirical estimate of $\Gamma_k(t)$. This is detailed in \Cref{alg:cap}, provided below.

\paragraph{Predictive distributions.}
In practice, the predictive distribution $q_f(Y_t\mid Z_{t-1},X_{t-1})$ is obtained directly from the model's output parameterisation. For classification tasks, this distribution is explicit: the network's logits define a categorical (or Bernoulli) distribution over discrete classes. A similar interpretation applies to generative models, where the predictive distribution corresponds to the model's generative distribution over data samples.

For regression tasks, the predictive distribution is determined by the likelihood implied by the training loss. For example, when optimising a mean-squared-error objective, the corresponding likelihood is Gaussian, so the model's prediction corresponds to the mean of a normal distribution. The predictive variance can be estimated empirically from the residual error distribution on a held-out validation set. Thus, even though most practical neural networks output only point predictions, their training objective implicitly defines a predictive distribution. We use this to evaluate the simulated marginalisation and empirically quantify forgetting.
\begin{figure}[H]\label{fig:pred_dist_example}
    \centering
    \includegraphics[width=\linewidth]{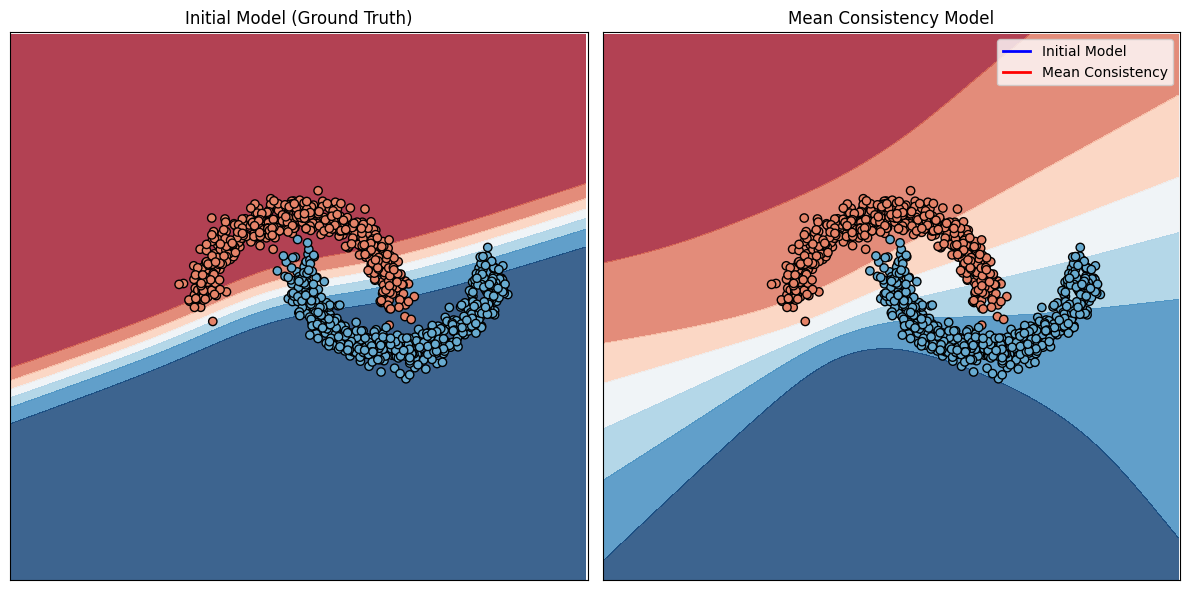}
    \caption{\textbf{Comparison of initial and 40-step simulated marginalisation.} Axes show the input space around the two-moon binary classification task. Example datapoints are overlaid and coloured by target class ($\text{red}=0$, $\text{blue}=1$). The background grid is shaded according to the classifier's logit values, from red (0) to blue (1), with white indicating uncertainty at 0.5. \textit{Left:} The initial predictive distribution $q(H^{t+k:\infty}\mid Z_{t-1},H_{0:t-1})$ serves as the baseline for consistent updates. \textit{Right:} The $k=40$ simulated marginalisation $q^*_k(H^{t+k:\infty}\mid Z_{t-1}, H_{0:t-1})$, showing how the predictive model is likely to evolve over 40 steps. The divergence in \Cref{def:forgetfulness} is computed between these two distributions.}
    \label{fig:classification_consistency_example}
\end{figure}

\paragraph{Divergence measures.}
The choice of divergence measure depends on the form of the predictive distribution.
In regression tasks with Gaussian likelihoods, the KL divergence between Gaussians yields a closed-form expression that can be empirically evaluated.
In classification tasks, outputs define categorical distributions over classes; therefore, the KL divergence (or cross-entropy) is also appropriate.
In generative settings, sample-based measures, like the MMD with a radial basis function (RBF) kernel, are used to compare distributions.
The same reasoning can be used in other tasks (e.g.\ RL), with the selected divergence measure providing a quantitative measure of how much the learner's predictive distribution has changed after simulated updates, returning the propensity to forget.

\begin{algorithm}[t]
  \caption{Computing the propensity to forget of a learner in a supervised classification setting.}\label{alg:cap}
  \begin{algorithmic}
    \State \textbf{Input:} Learner with state $Z_{t-1}$, update function $u$, inputs $X_{t:\infty}$, \# of particles $N$, horizon $k$, history $H_{0:t-1}$
    \State Compute initial predictive distribution $q(Y_{t+k:\infty}\mid Z_{t-1},X_{t+k-1:\infty})$
    \State Initialise $N$ particle copies of the learner: $\{Z_{t-1}^{(n)}\}_{n=1}^N$
    \For{each particle $n=1$ to $N$}
        \For{$s=1$ to $k$}
            \State Sample target $Y_{t+s}^{(n)}\sim q_f(\cdot\mid Z_{t+s-1}^{(n)},X_{t+s-1}^{(n)})$
            \State Sample input $X_{t+s}^{(n)}\sim p_e(\cdot\mid H_{0:t+s-1}, Y_{t+s}^{(n)})$ \TriComment{Component not modelled. Sample from validation set.}
            \State Update particle state: $Z_{t+s}^{(n)}\sim u(\cdot\mid Z_{t+s-1}^{(n)},X_{t+s}^{(n)},Y_{t+s}^{(n)})$
        \EndFor
        \State Compute predictive distribution for future inputs $q_k^{(n)}(Y_{t+k:\infty}\mid Z_{t+k-1}^{(n)},X_{t+k-1:\infty}^{(n)})$
    \EndFor
    \State Form a mixture of the $N$ particle predictions $q_k^*(Y_{t+k:\infty}\mid Z_{t+k-1},X_{t+k-1:\infty})$
    \State Compute $\Gamma_k(t)=\mathrm D_{\mathrm{KL}}\left( q(Y_{t+k:\infty}\mid Z_{t-1},X_{t+k-1:\infty}) \,\|\, q_k^*(Y_{t+k:\infty}\mid Z_{t+k-1},X_{t+k-1:\infty}) \right)$
    \State {\bfseries Return:} $\Gamma_k(t)$
  \end{algorithmic}
\end{algorithm}

\paragraph{Empirical implementation of the simulated marginalisation.}
Computing $\Gamma_k(t)$ in practice requires that the simulated future updates \emph{do not interfere} with the model's actual training dynamics.
Across all tasks, this involves creating independent Monte Carlo replicas (particles) of the current model and propagating each replica through $k$ simulated update steps.
These updates use held-out validation data or simulated interactions in an environment if the learner does not model all components required for a simulated marginalisation.
This produces a set of post-update models whose predictive distributions can be aggregated to approximate the $k$-step marginalised distribution $q_k^*$.
All training, sampling, and updates performed on these particles are independent of each other and the main training loop.

This is an intricate procedure in RL, where each particle maintains a separate copy of the Q-network, target network, optimiser, and replay buffer.
Particles are propagated through $k$ steps of environment interactions, with the resulting predictive distributions used to measure $\Gamma_k(t)$ with some appropriate divergence measure.

Overall, this empirical approach generalises across supervised, generative, and RL settings.
\begin{itemize}[left=0pt,nosep]
  \item Simulate future updates independently,
  \item Aggregate particle predictive distributions to form the $k$-step predictive distribution,
  \item Measure the divergence between the initial predictive distribution and the $k$-step predictive distribution.
\end{itemize}

\paragraph{Visualising the propensity to forget}
\Cref{fig:classification_consistency_example} compares a single-layer neural network's futures distribution at the current training step with the same model but after $40$ additional updates on futures data. This allows us to visualise the divergence between the predictive distribution before and after a simulated marginalisation. This difference quantifies the model's propensity to forget, as defined in \Cref{def:forgetfulness}.

\subsection{Supervised classification example}\label{app:classification_example}
Consider a supervised classification setting with a neural network learner. To approximate future inputs $X_{t:\infty}$, we sample uniformly over the empirical distribution of inputs observed thus far. This is acquired by sampling $k$, $X$ values from a held-out validation set, $X_{t:t+k-1}$, and using the remaining inputs to evaluate predictive distributions, approximating $X_{t+k:\infty}$.

We first compute the predictive distribution of the current model, $Z_{t-1}$.
\begin{equation}
    q(Y_{t+k:\infty}\mid Z_{t-1},X_{t+k-1:\infty}),
\end{equation}
which will act as the reference distribution in the KL divergence.

To form the $k$-step simulated marginalisation, we perform a Monte Carlo estimate: make $N$ independent particle copies of the learner and, for each particle $n$ and for $s=1,\dots,k$, we do the following:
\begin{align}
\begin{aligned}
    &\text{(sample target)} && Y_{t+s}^{(n)}\sim q_f(\cdot\mid Z_{t+s-1}^{(n)},X_{t+s-1}^{(n)}), \\
    &\text{(sample input)} && X_{t+s}^{(n)}\sim q_e(\cdot\mid H_{0:t+s-1}, Y_{t+s}^{(n)}), \\
    &\text{(update state)} && Z_{t+s}^{(n)}\sim u(\cdot\mid Z_{t+s-1}^{(n)},X_{t+s}^{(n)},Y_{t+s}^{(n)}).
\end{aligned}
\end{align}
After $k$ steps, each particle yields a predictive distribution over the future inputs,
\begin{equation}
    q_k^{(n)}(Y_{t+k:\infty}\mid Z_{t+k-1}^{(n)},X_{t+k-1:\infty}^{(n)}).
\end{equation}
We form the uniform mixture to acquire the simulated marginalisation,
\begin{equation}
    q_k^*(Y_{t+k:\infty}\mid Z_{t+k-1},X_{t+k-1:\infty}).
\end{equation}
We can then compute the learner's propensity to forget:
\begin{equation}
    \Gamma_k(t)=D_{\mathrm{KL}}\left( q(Y_{t+k:\infty}\mid Z_{t-1},X_{t+k-1:\infty}) \,\|\, q_k^*(Y_{t+k:\infty}\mid Z_{t+k-1},X_{t+k-1:\infty}) \right).
\end{equation}
This process is described in \Cref{alg:cap}.

\paragraph{Computational considerations.}
The primary computational cost of this procedure stems from propagating $N$ independent particles over $k$ synthetic update steps, each involving a forward pass and parameter update of the learner. This can be significant for large neural networks or long consistency horizons $k$. Two sources of approximation error arise in practice. First, the predictive distributions $q_f$ and $q_k^*$ are evaluated only on a finite validation set, introducing sampling error that decreases with the number of held-out inputs. Second, the mixture $q_k^*$ is only a Monte Carlo approximation to the expectation over the learner's self-generated future. Both sources can be reduced by increasing the size of the validation set or the number of particles, though at the cost of increased computational load.

\vfill
\begin{center}
    --appendices continue on next page--
\end{center}

\clearpage

\section{Additional results}\label{app:results}
In this section, we discuss additional results, including Bayesian learners, permutation sensitivity, and an expanded discussion of \Cref{sec:empirical}.
\begin{figure}[ht]
    \centering
    \includegraphics[width=0.9\linewidth]{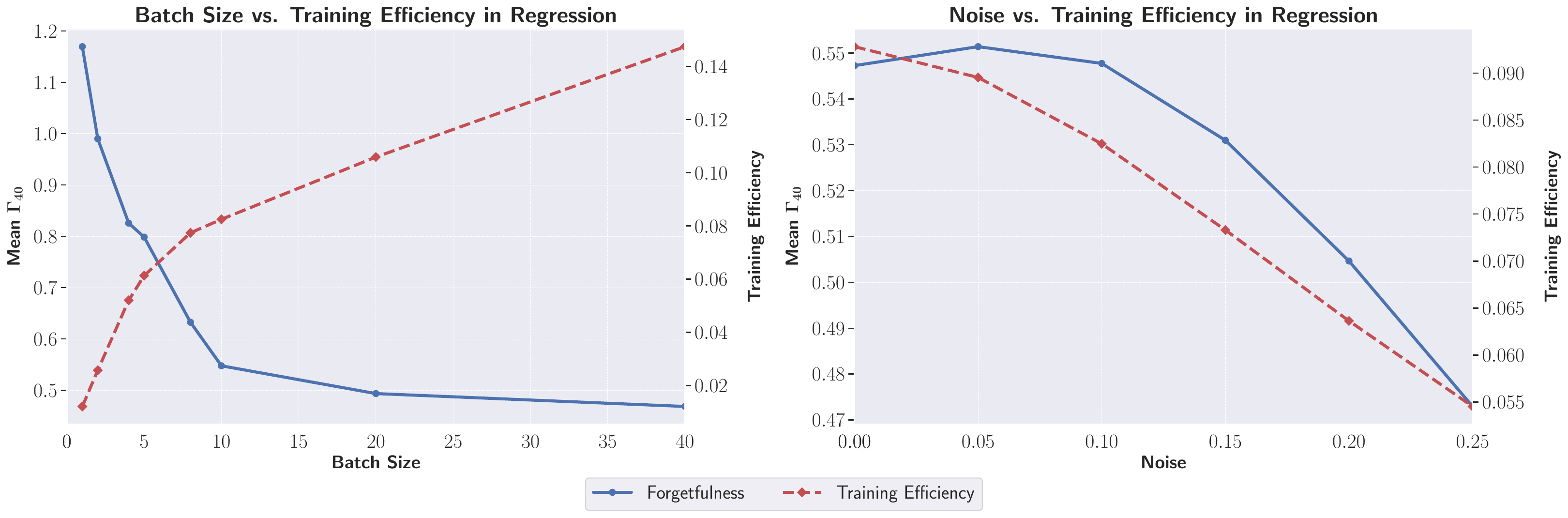}
    \caption{\textbf{Optimal forgetting is not necessarily zero.} We examine how a regression learner's training efficiency and propensity to forget vary under different hyperparameter settings. \emph{Left:} Increasing the batch size reduces forgetting while improving training efficiency. Forgetting plateaus once the batch size approaches the dataset size (40 datapoints), at which point training efficiency increases with minimal further reduction in forgetting. \emph{Right:} Reducing the noise in the dataset leads to substantially higher training efficiency; however, in low-noise regimes, the learner becomes more sensitive to self-consistency perturbations and therefore forgets more despite performing better on the task.}
    \label{fig:training_eff}
\end{figure}

\subsection{Forgetting in supervised learning}\label{app:forgetting_supervised}
Here we expand on the empirical results presented in \S\ref{sec:empirical}, examining how the dynamics of forgetting vary with model architecture, hyperparameters, and optimisation settings. In addition to the training efficiency analysis in \Cref{fig:training_efficiency_il_forgetting}, we observe that hyperparameters can simultaneously shape both training efficiency and forgetting in \Cref{fig:training_eff}.
\begin{figure}[H]
    \centering
    \includegraphics[width=0.9\linewidth]{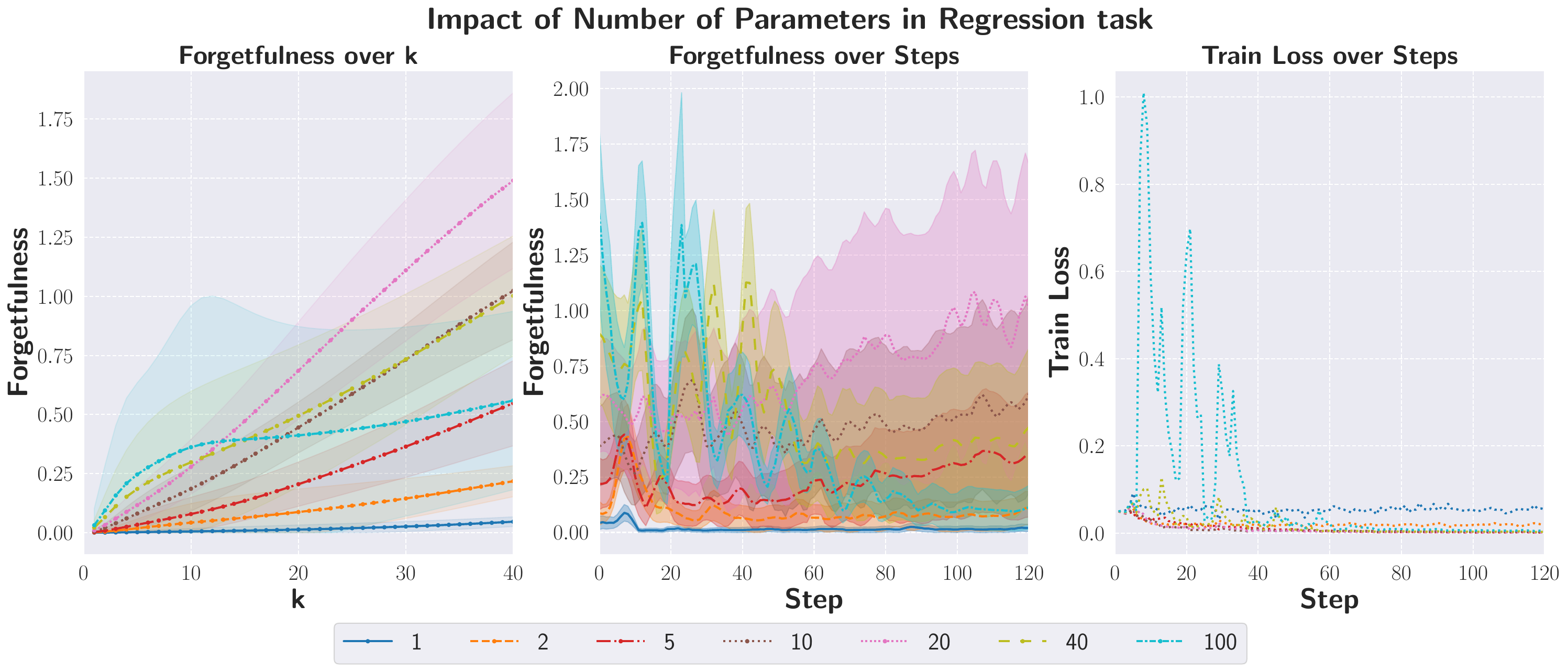}
    \caption{\textbf{Impact of model size on forgetting dynamics.} Plots illustrating the impact of varying numbers of hidden-layer parameters in a single-hidden-layer neural network on a regression task with 40 training datapoints (details in \Cref{app:implementation}). \textit{Left:} Forgetfulness as a function of the number of updates $k$, showing how the learner's propensity to forget evolves over update steps. \textit{Middle:} Forgetfulness throughout training, highlighting changes in the learner's propensity to forget over time. \textit{Right:} Training loss curves, comparing learning dynamics across model sizes. We observe that forgetting dynamics are strongly influenced by model size: forgetting increases with model size until the learner's parameter count approaches or exceeds the task's effective size, after which it declines again. These dynamics have a major impact on training efficiency, as shown in \Cref{fig:training_efficiency_il_forgetting}.}
    \label{fig:regression-capacity}
\end{figure}
\paragraph{Effect of model size.}
\Cref{fig:regression-capacity} examines how model size affects forgetting and learning efficiency on the regression task detailed in \Cref{app:implementation}. Each plot compares networks with different hidden-layer sizes, trained on 40 data points spanning both under- and over-parameterised tasks. We find that forgetting generally increases with model size, peaking near the point where the number of parameters matches the task's effective complexity, and then decreases again in the highly overparameterised task. This inverted trend mirrors the behaviour of training efficiency discussed in \Cref{fig:training_efficiency_il_forgetting}.

\begin{figure}[H]
    \centering
    \includegraphics[width=0.9\linewidth]{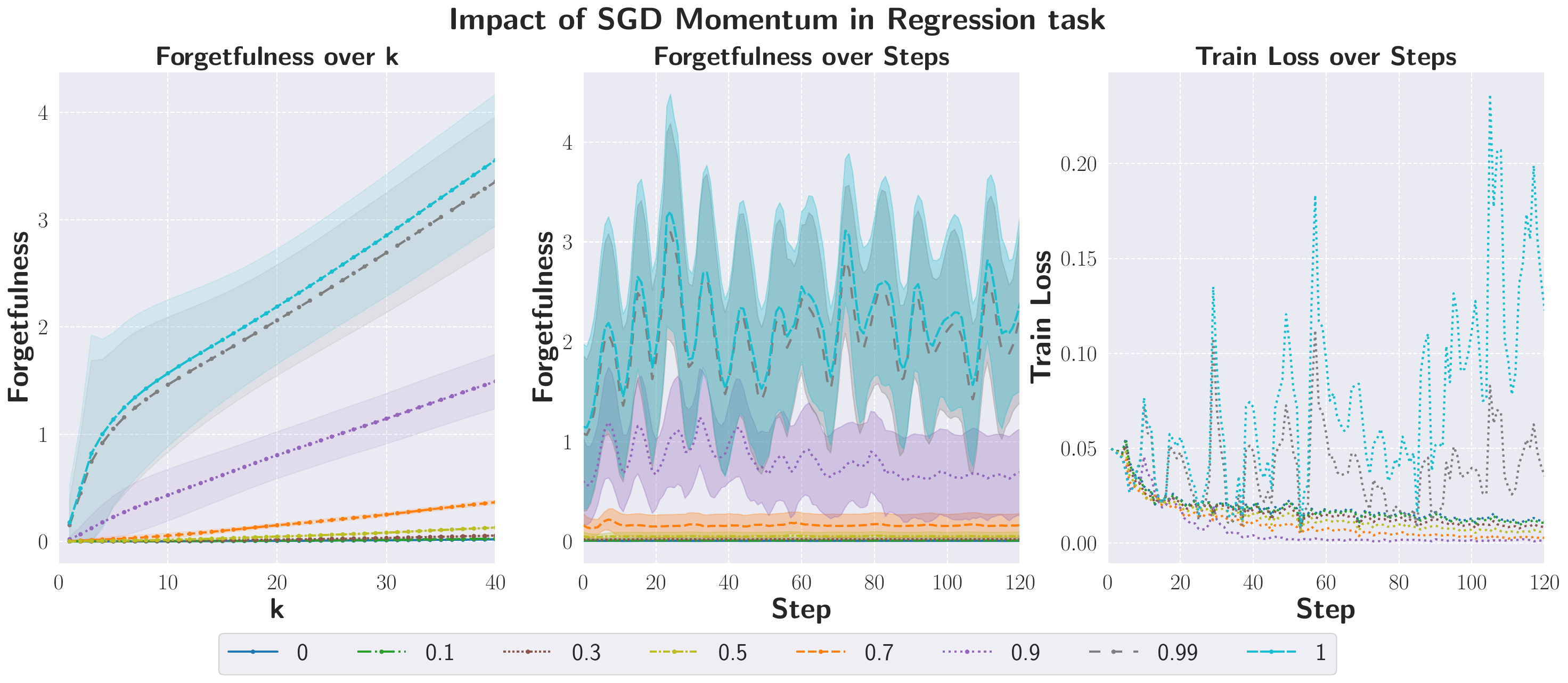}
    \caption{\textbf{Effect of the momentum parameter on forgetting dynamics.} Plots illustrating the impact of varying momentum coefficients during training of a single-hidden-layer neural network on the sinusoid regression task. \textit{Left:} Forgetfulness as a function of the number of updates $k$. \textit{Middle:} Forgetfulness over the course of training. \textit{Right:} Training loss curves for each momentum value. We observe that forgetting dynamics exhibit periodicity across updates, with the mean forgetting and the oscillation amplitude increasing as momentum increases. Optimal training efficiency occurs at a momentum of 0.9, corresponding to an optimal balance of adaptation and stability.}
    \label{fig:regression-momentum}
  \vspace{-1em}
\end{figure}
\paragraph{Effect of momentum.}
In \Cref{fig:regression-momentum}, we vary the momentum coefficient in stochastic gradient descent while training on a regression task. The propensity to forget exhibits periodic fluctuations that increase with higher momentum values. Both mean forgetting, and the oscillatory behaviour grows with momentum, indicating a trade-off between stability and adaptivity. The most efficient training occurs when moderate forgetting coincides with stable, fast convergence.

\begin{figure}[H]
    \centering
    \includegraphics[width=0.9\linewidth]{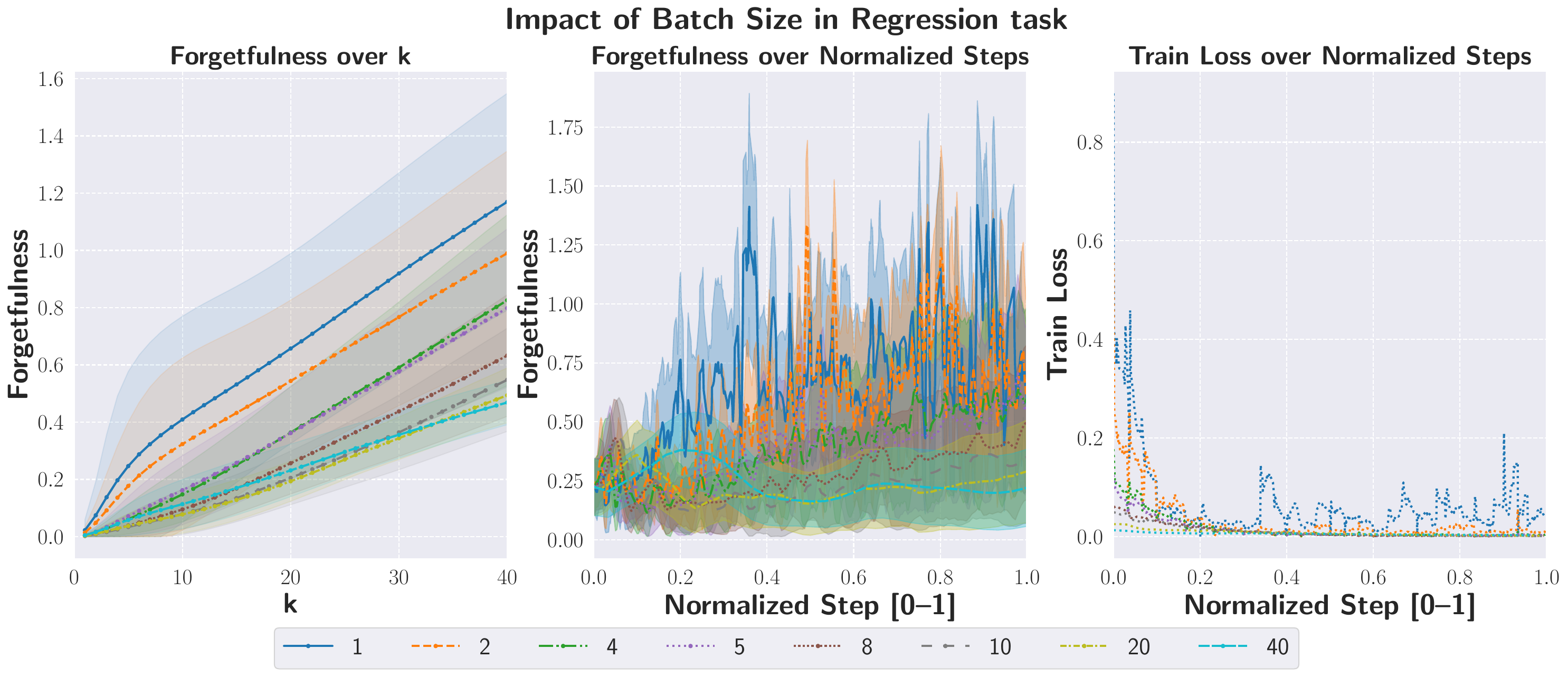}
    \caption{\textbf{Effect of batch size on forgetting dynamics.} Lines and markers indicate different batch sizes used in training a single-hidden-layer neural network on the sinusoid regression task. \textit{Left:} Forgetfulness as a function of the number of updates $k$. \textit{Middle:} Forgetfulness over the course of training. \textit{Right:} Training loss curves across batch sizes. Smaller batches exhibit significant fluctuations and high variability, resulting in unstable learning and reduced efficiency. Increasing batch size stabilises learning and reduces forgetting, until the effect plateaus at a batch size of 10, beyond which forgetting remains non-zero but is approximately the same across batch sizes.}
    \label{fig:regression-batch_size}
  \vspace{-1em}
\end{figure}
\paragraph{Effect of batch size.}
\Cref{fig:regression-batch_size} studies the influence of batch size on forgetting dynamics. Smaller minibatches lead to higher update variability, resulting in larger oscillations and a greater propensity to forget as training progresses. As batch size increases, efficiency increases and forgetting decreases, plateauing at around one-quarter of the training dataset size. Beyond this, forgetting remains non-zero and is very similar across batch sizes greater than one-quarter of the training dataset.

\begin{figure}[ht]
    \centering
    \includegraphics[width=\linewidth]{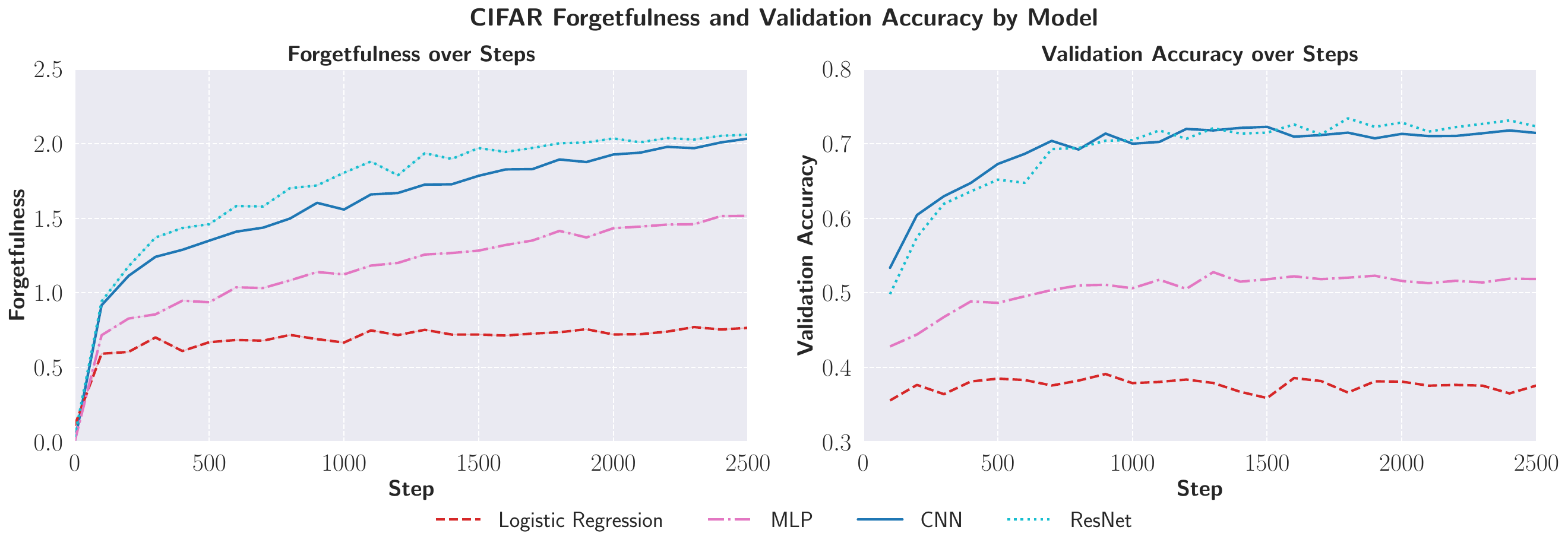}
    \caption{\looseness=-1\textbf{Forgetting dynamics across model architectures on a high-dimensional classification task.} Different-coloured lines and marker styles denote different model types: logistic regression, MLP, CNN, and ResNet. \textit{Left:} Forgetfulness over training updates. \textit{Right:} Validation accuracy over training time. While CNNs and ResNets exhibit substantially higher levels of forgetting than logistic regression and MLPs (with ResNets forgetting the most), they achieve the best task performance. This implies that the most effective deep learning models do not necessarily minimise forgetting.}
    \label{fig:cifar-forgetfulness}
\end{figure}
\paragraph{Effect of architecture.}
Finally, \Cref{fig:cifar-forgetfulness} examines how model architecture influences forgetting on a high-dimensional classification task (CIFAR-10). We compare logistic regression, a multilayer perceptron (MLP), a convolutional neural network (CNN), and a ResNet. CNNs and ResNets exhibit considerably higher forgetting than logistic regression and MLPs, yet they achieve better task performance. This indicates that the most effective deep learning models do not necessarily minimise forgetting; instead, they maintain a balance where moderate forgetting supports continued adaptation and improved generalisation.

\paragraph{Summary.}
Together, these results (\Cref{fig:training_eff}, \Cref{fig:regression-capacity}, \Cref{fig:regression-momentum}, and \Cref{fig:regression-batch_size}) show that forgetting is a pervasive property of deep learning systems, shaped by both the update dynamics and the learner's hyperparameters. Consistent with \Cref{fig:training_efficiency_il_forgetting}, regimes with improved efficiency do not necessarily minimise forgetting. While high levels of forgetting can destabilise training, moderate levels appear beneficial: enabling adaptability while preserving some past information. The relationship between forgetting and learning efficiency is thus nuanced: \emph{efficient learning requires not the absence of forgetting but its regulation}.

\clearpage
\subsection{Forgetting in reinforcement learning}\label{app:forgetting_rl}
Here, we discuss the impact of hyperparameters in RL on forgetting dynamics.

\paragraph{Buffer size.}
The replay buffer size determines which experiences are retained and sampled during Q-learning updates. A small buffer restricts the effective training distribution to the most recent transitions, while a large buffer maintains long-range temporal support.

\begin{figure}[ht]
    \centering
    \begin{subfigure}{0.8\linewidth}
        \centering
        \includegraphics[width=\linewidth]{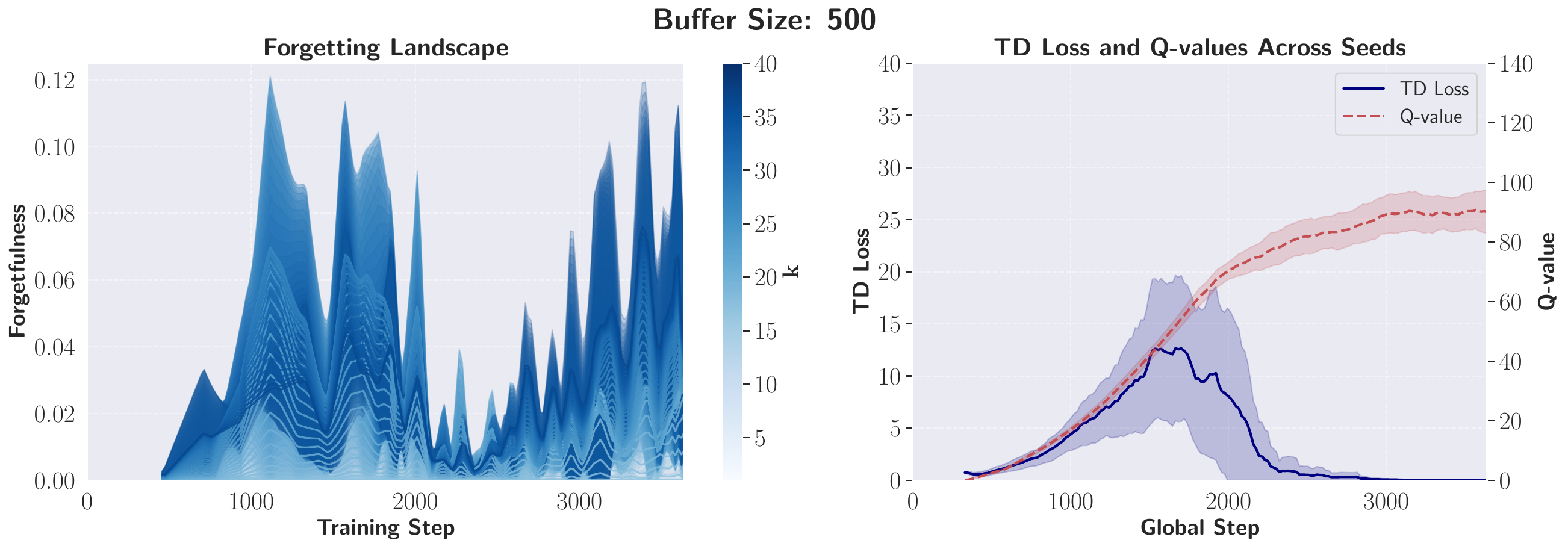}
    \end{subfigure}
    \begin{subfigure}{0.8\linewidth}
        \centering
        \includegraphics[width=\linewidth]{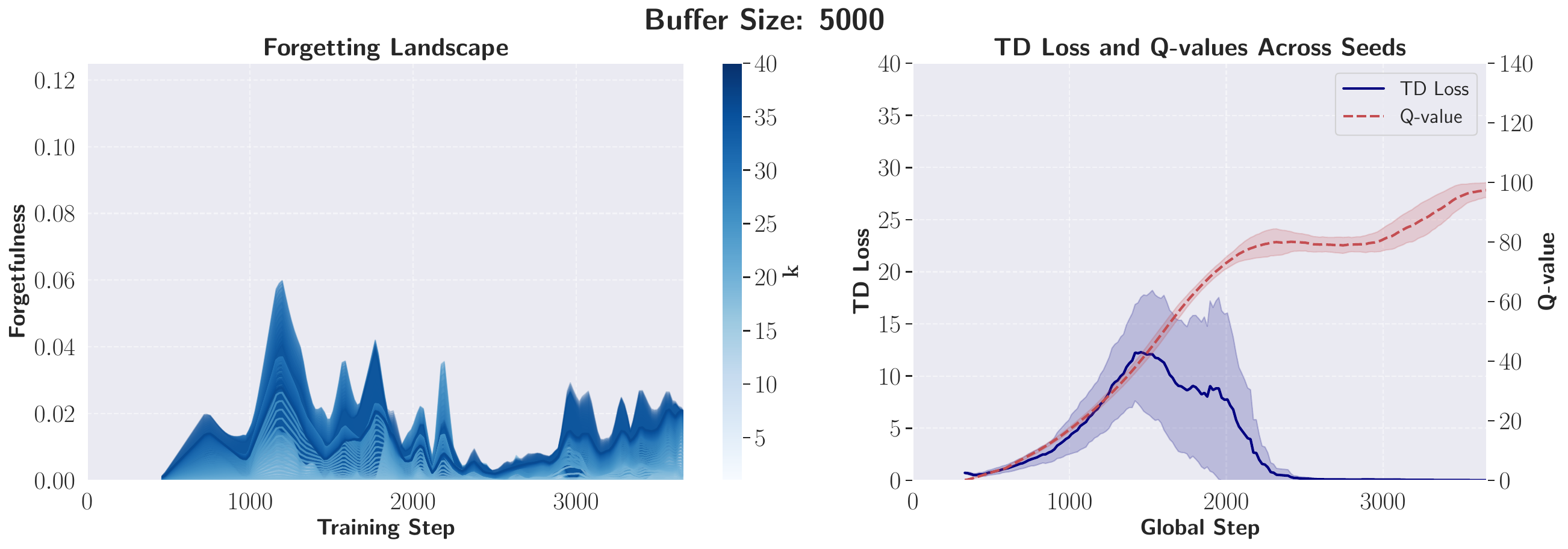}
    \end{subfigure}
    \begin{subfigure}{0.8\linewidth}
        \centering
        \includegraphics[width=\linewidth]{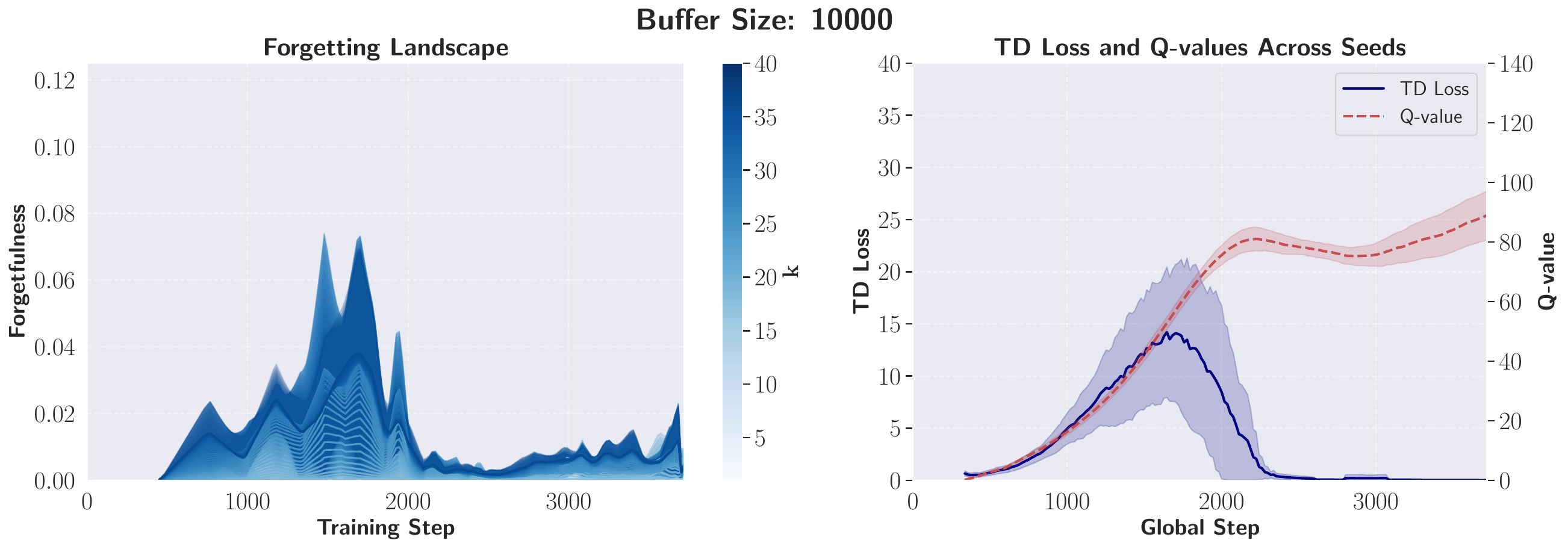}
    \end{subfigure}
    \caption{\textbf{Replay buffers regulate prediction support.} Forgetting landscapes for different replay buffer sizes. Small buffers produce high, unstable forgetting because the learner continually overwrites supported predictions as the data distribution shifts. Larger buffers stabilise the distribution and reduce forgetting, but overly large buffers reintroduce outdated transitions, which can break self-consistency, leading to moderate forgetting and reduced learning efficiency.}
    \label{fig:rl-buffer_sizes}
\end{figure}
With very small buffers, the training distribution shifts rapidly as old transitions are discarded. The learner, therefore, repeatedly overwrites previously supported predictions, resulting in large, unstable forgetting dynamics. As the buffer size increases, forgetting decreases and becomes more stable: predictions are supported by a more stable training distribution, so self-consistency violations decrease. However, extremely large buffers also reintroduce unsupported transitions, causing the learner to train on outdated targets and leading to inconsistency. Thus, both forgetting and task performance are optimised at intermediate buffer sizes. This is observed in \Cref{fig:rl-buffer_sizes}.

\paragraph{Target network update rate.}
$\tau$ controls the target Q-network update rate. Therefore, $\tau$ controls how quickly the support for predictions evolves. 
\begin{figure}[H]
    \centering
    \begin{subfigure}{0.8\linewidth}
        \centering
        \includegraphics[width=\linewidth]{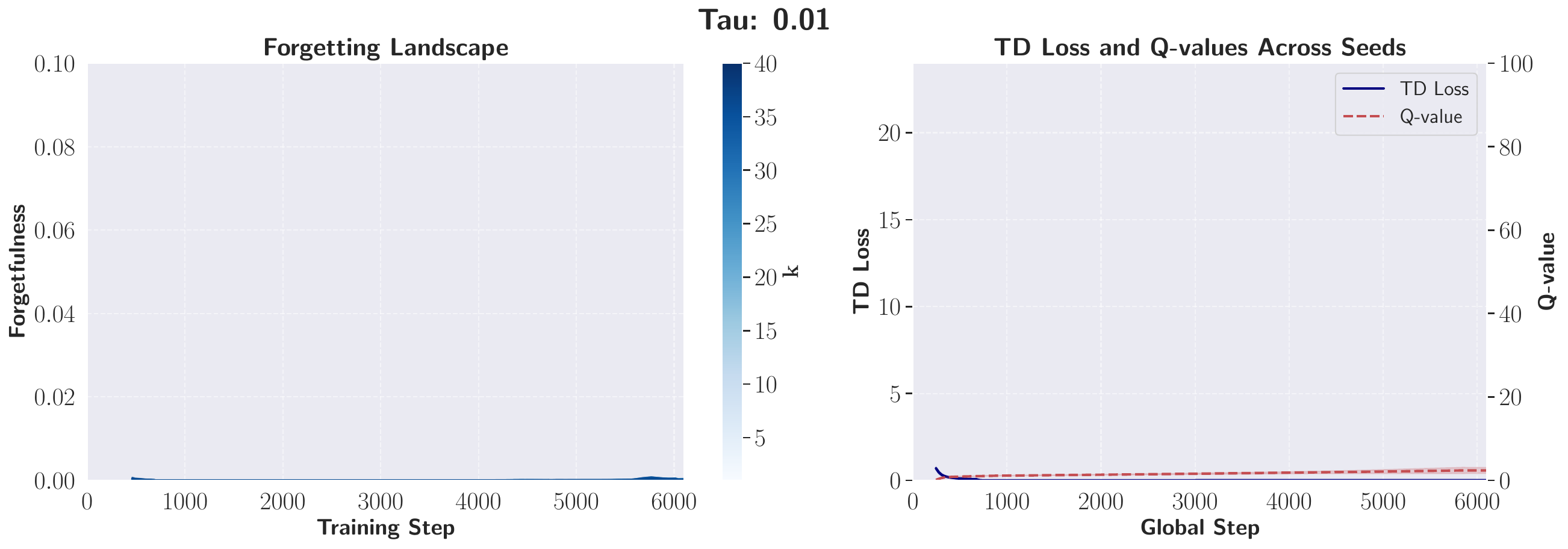}
    \end{subfigure}
    \begin{subfigure}{0.8\linewidth}
        \centering
        \includegraphics[width=\linewidth]{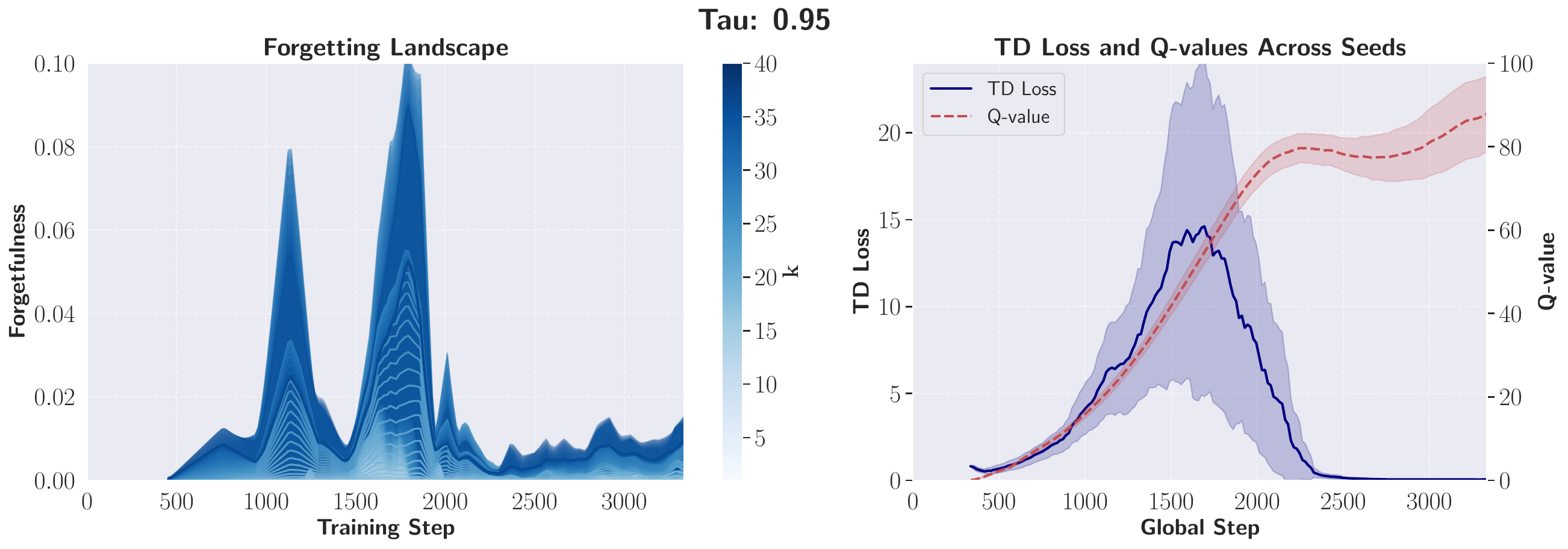}
    \end{subfigure}
    \begin{subfigure}{0.8\linewidth}
        \centering
        \includegraphics[width=\linewidth]{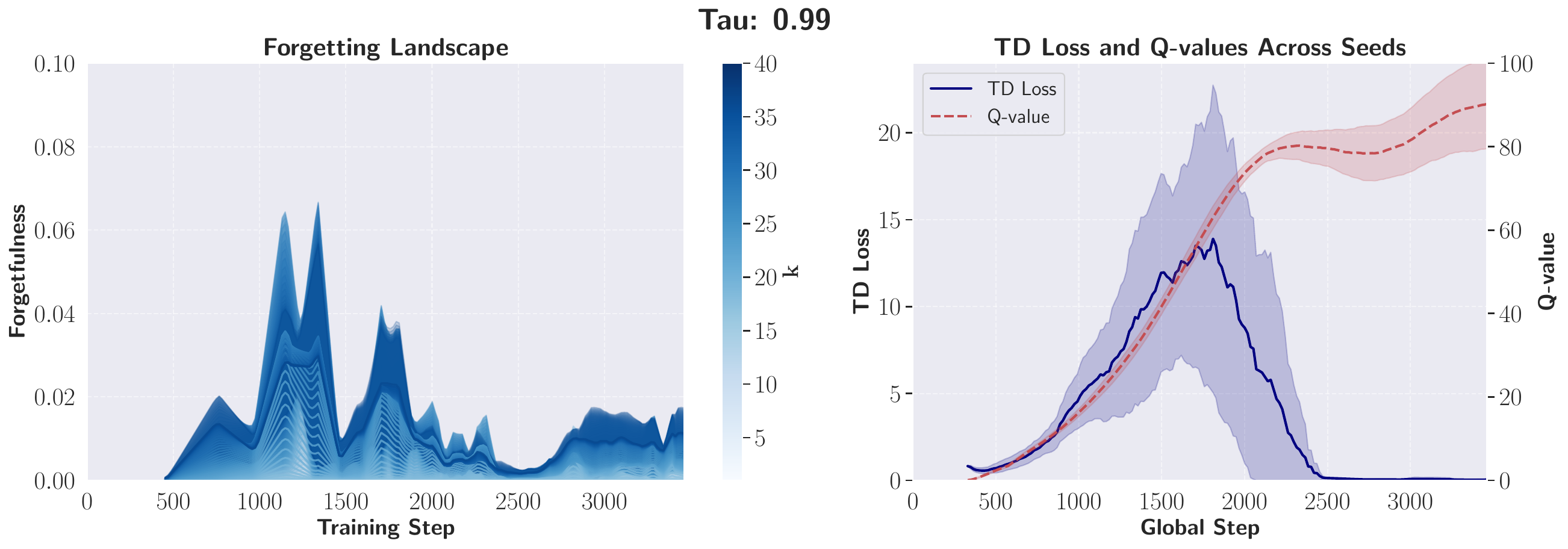}
    \end{subfigure}
    \begin{subfigure}{0.8\linewidth}
        \centering
        \includegraphics[width=\linewidth]{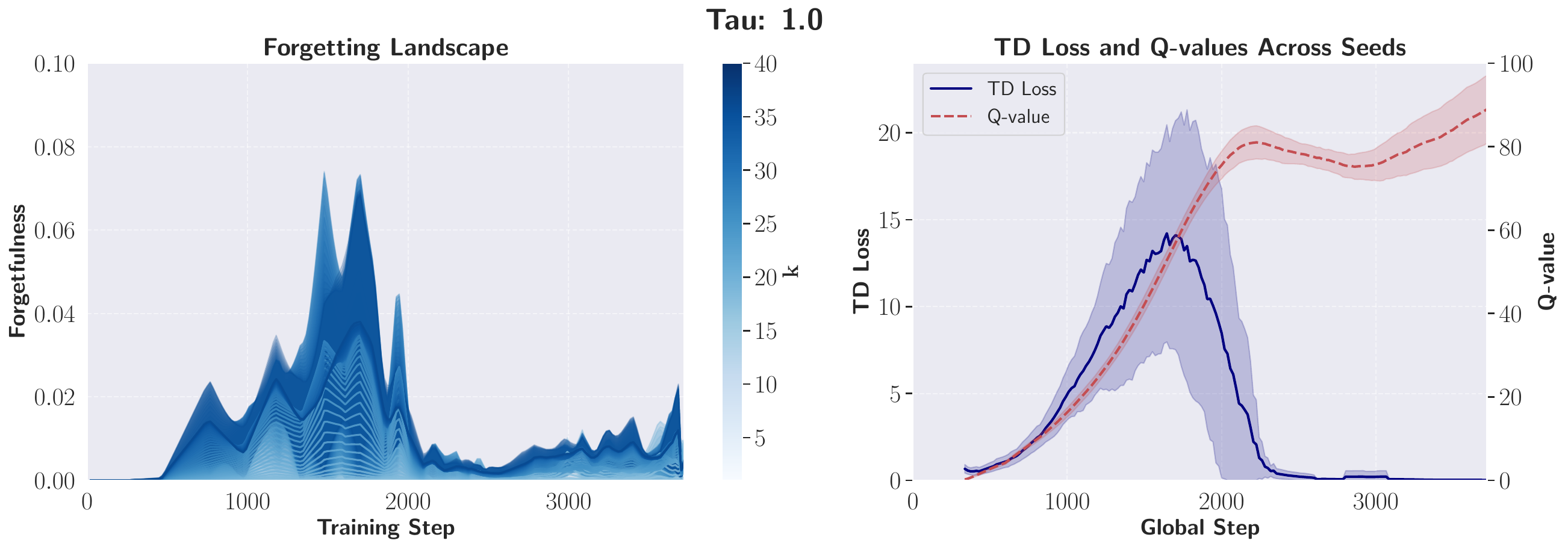}
    \end{subfigure}
    \caption{\textbf{Target updates trade off stability and adaptability.} Forgetting landscapes for different target network update rates $\tau$. Large $\tau$ yields less-stable forgetting dynamics due to rapidly shifting targets; small $\tau$ yields smooth but uninformative updates that prevent learning. Intermediate $\tau$ values best maintain support for predictions and thus yield balanced forgetting and effective learning.}
    \label{fig:rl-tau}
\end{figure}

When $\tau$ is too large, the target network is updated frequently, giving the learner a less stable reference. Predictions can become unsupported more frequently, potentially producing higher, less stable forgetting dynamics. When $\tau$ is very small, the target network is rarely updated. These result in fixed stable targets; however, they also impair the learner's ability to learn. The learner may not be forgetting, but they are also failing to learn. When $\tau$ is moderate, the target network adjusts gradually, allowing predictions at time $t$ to remain approximately supported by subsequent updates, which is important for self-consistency.

Consequently, both forgetting and performance show optimal behaviour at intermediate $\tau$: the target evolves slowly enough to provide support for predictions, yet quickly enough to adapt the fixed point toward a self-consistent solution (see \Cref{fig:rl-tau}).

\paragraph{Training frequency.}
Training frequency determines how often the learner applies updates in response to environmental interactions. This determines the rate at which the training distribution shifts.
\begin{figure}[ht]
    \centering
    \begin{subfigure}{0.8\linewidth}
        \centering
        \includegraphics[width=\linewidth]{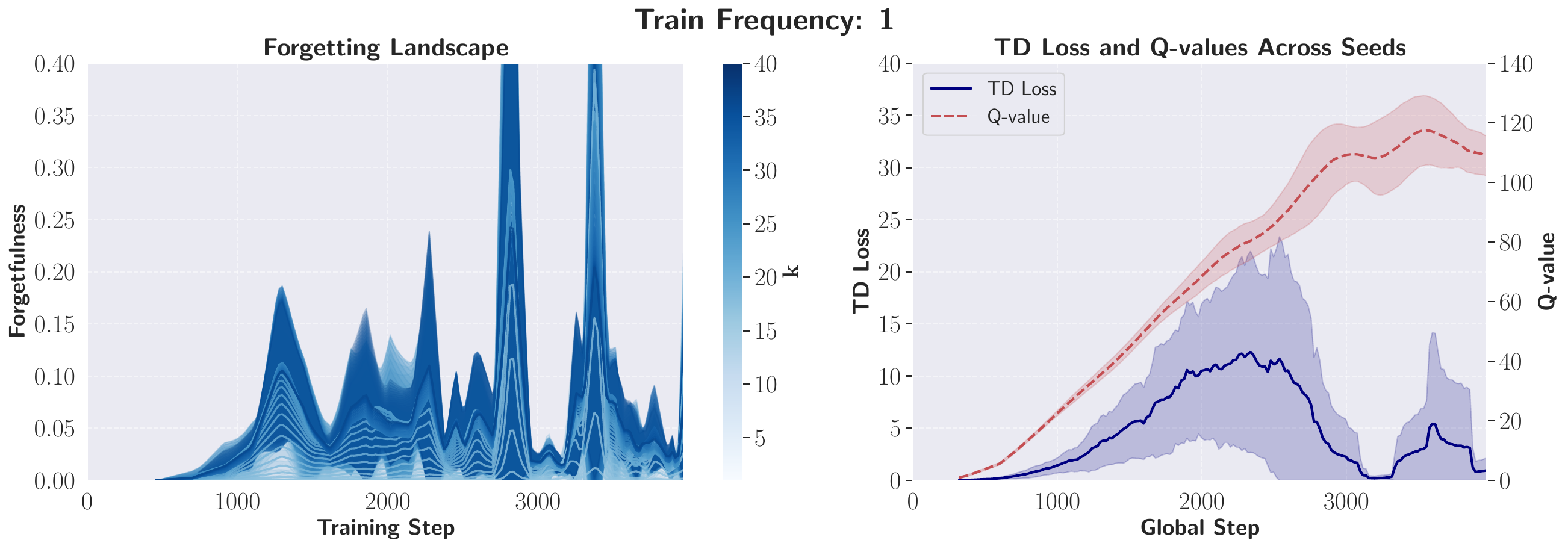}
    \end{subfigure}
    \begin{subfigure}{0.8\linewidth}
        \centering
        \includegraphics[width=\linewidth]{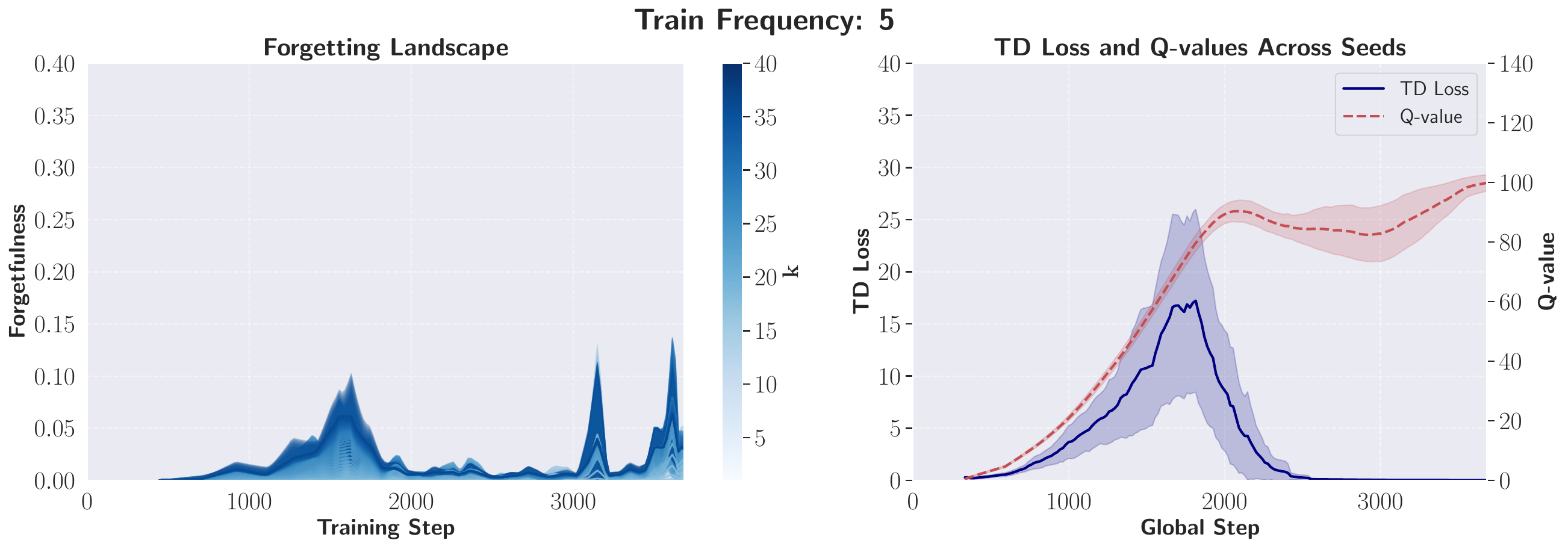}
    \end{subfigure}
    \begin{subfigure}{0.8\linewidth}
        \centering
        \includegraphics[width=\linewidth]{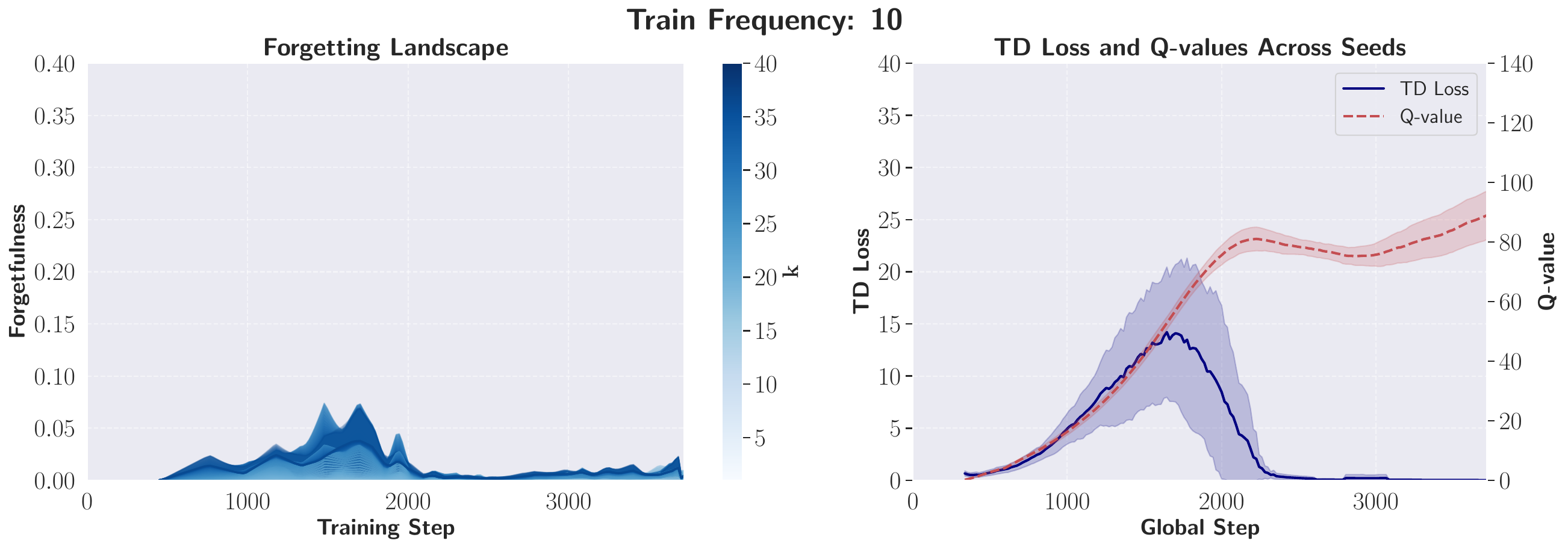}
    \end{subfigure}
    \begin{subfigure}{0.8\linewidth}
        \centering
        \includegraphics[width=\linewidth]{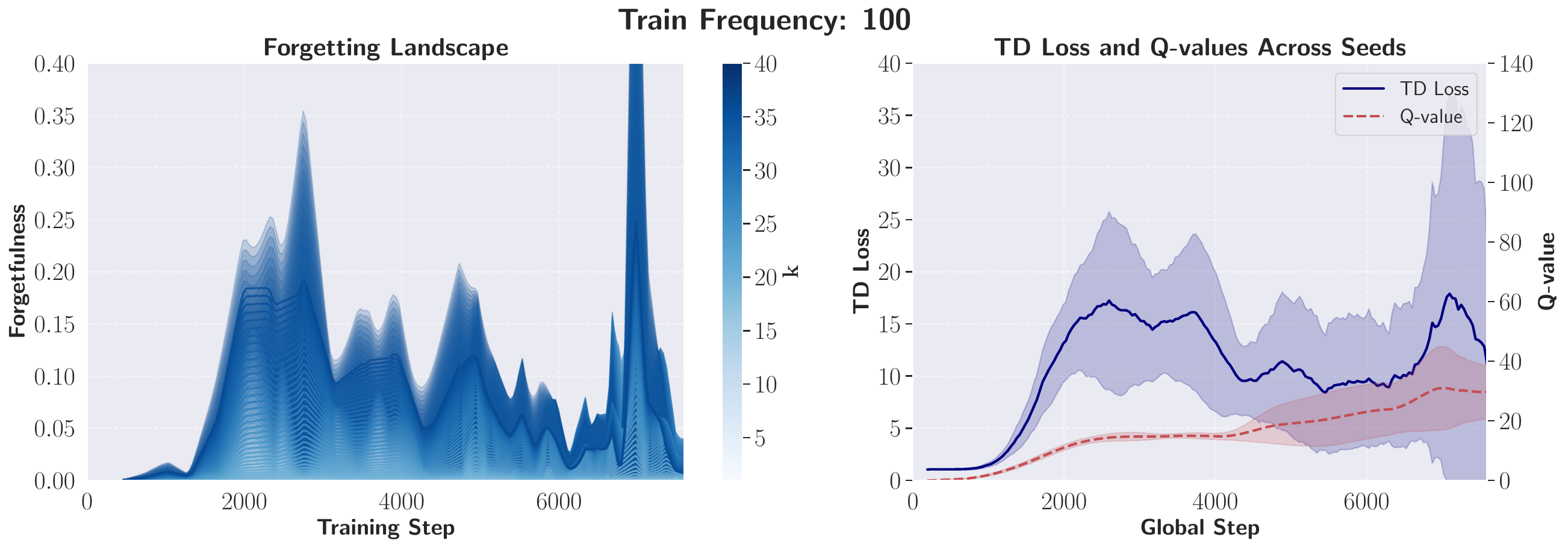}
    \end{subfigure}
    \caption{\textbf{Training frequency determines forgetting.} Forgetting landscapes for different training frequencies. Updating too frequently causes highly unstable forgetting due to rapid shifts in the effective training distribution. Updating too infrequently also causes large shifts in the effective training distribution. A trade-off must be found to achieve more self-consistent updates.}
    \label{fig:rl-train_freqs}
\end{figure}

When the learner makes frequent updates (low training-frequency values), the effective training distribution shifts frequently. As such, updates are frequently performed on shifting distributions, leading to inconsistent predictive distributions over time.

When the learner makes moderately frequent updates, the effective training distribution is quite stable because the learner accumulates small batches of experience before updating. This creates smaller, more stable forgetting because predictions have more persistent support: the effective training distribution changes gradually, and model updates are less likely to be inconsistent.

However, when the learner makes infrequent updates, the effective training distribution again shifts frequently: the learner is updated so infrequently that each update incorporates a highly non-stationary batch of transitions. This again causes model updates to be inconsistent, again breaking support for many of the predictions the learner made earlier, and the forgetting dynamics again increase and become more unstable. In this setting, learning performance also decreases because update frequency affects the learner's ability to learn.

Thus, optimal self-consistency emerges only at moderate training frequencies (see \Cref{fig:rl-train_freqs}).

\paragraph{Summary.}
Across our RL experiments, we observe that forgetting dynamics are strongly influenced by hyperparameters such as buffer size, target network update rate, $(\tau)$, and training frequency. Smaller buffer sizes lead to high and unstable forgetting due to insufficient replay support; very low $\tau$ values produce stable forgetting dynamics but prevent meaningful learning; and both low and high training frequencies induce chaotic forgetting dynamics. In all cases, these dynamics, previously supported by the learner's state, are no longer reinforced when support is insufficient, leading to greater forgetting.

We also note that forgetting tends to track the TD loss throughout training, suggesting that temporal-difference updates implicitly regulate forgetting dynamics. These results highlight that managing forgetting is not just a consequence of network design or training heuristics, but a \emph{vital consideration in the design of RL algorithms}: poor control of forgetting leads to unstable learning, lower sample efficiency, and suboptimal performance. Optimal algorithm design, therefore, requires balancing the reinforcement of past knowledge with the acquisition of new information.
\clearpage

\vfill
\begin{center}
    --appendices continue on next page--
\end{center}

\section{Experiment details}\label{app:implementation}
We used NVIDIA GeForce RTX 2080 Ti GPUs to run our experiments.
\Cref{tab:hyperparameters,tab:hyperparameters1} lists all of the hyperparameters that were used in the experiments.
Upon acceptance, our code will be made available on GitHub.

\begin{table}[H]
\centering
\caption{Hyperparameters for generative modelling and DQN reinforcement learning experiments.}
\begin{tabularx}{\linewidth}{lXl}
\toprule
\textbf{Hyperparameter} & \textbf{Description} & \textbf{Values / Type} \\
\midrule

\multicolumn{3}{l}{\textbf{Propensity to Forget, $\Gamma_k(t)$, Settings}} \\
k & Number of forgetfulness updates & 40 \\
num\_particles & Number of particles for Monte Carlo approximation & 1000 \\
\midrule

\multicolumn{3}{l}{\textbf{Class Incremental-Learning} (Two Moons)} \\
noise & Dataset noise & 0.1 \\
epochs & Number of training epochs & 30 \\
num\_tasks & Number of tasks & 2 \\
batch\_size & Batch size & 25 \\
num\_samples & Number of training samples per task & 100 \\
num\_val\_samples & Number of validation samples & 100 \\
lr & Learning rate & 0.1 \\
optimiser & Optimiser & Adam \\
hidden\_dim & Hidden layer size & 10 \\
\midrule

\multicolumn{3}{l}{\textbf{Reinforcement Learning, DQN} (CartPole)} \\
batch\_size & Number of transitions sampled per gradient update & 128 \\
buffer\_size & Replay buffer size & 10,000 \\
start\_e & Initial exploration rate ($\epsilon$) & 1.0 \\
end\_e & Final exploration rate ($\epsilon$) & 0.05 \\
exploration\_fraction & Fraction of training during which $\epsilon$ decays & 0.5 \\
learning\_starts & Number of steps before training starts & 10,000 \\
train\_frequency & Number of steps between gradient updates & 10 \\
target\_network\_frequency & Frequency of hard target network updates & 500 \\
tau & Soft target update rate & 1.0 \\
gamma & Discount factor for future rewards & 0.99 \\
lr & Learning rate for optimiser & 0.00025 \\
optimiser & Optimiser used for training & Adam \\
num\_eval\_steps & Number of steps per evaluation & 1,000 \\
total\_timesteps & Total number of environment interactions & 200,000 \\
num\_parameters & Number of model parameters & 5 \\
Divergence measure & Softmax raw logits to produce a categorical distribution over actions & KL divergence \\
\bottomrule
\end{tabularx}
\label{tab:hyperparameters1}
\end{table}

\begin{table}[ht]
\centering
\caption{Summary of hyperparameters and settings for all tasks.}
\begin{tabularx}{\linewidth}{lXl}
\toprule
\textbf{Hyperparameter} & \textbf{Description} & \textbf{Values / Type} \\
\midrule

\multicolumn{3}{l}{\textbf{Regression} (Sinusoid)} \\
noise & Observation noise & 0.1 \\
epochs & Number of training epochs & 30 \\
batch\_size & Batch size & 10 \\
num\_samples & Number of training samples & 40 \\
num\_val\_samples & Number of validation samples & 100 \\
lr & Learning rate & 0.1 \\
optimiser & Optimiser & Adam \\
num\_parameters & Number of model parameters & 5 \\
Divergence measure & The divergence measure used in this task & KL divergence, Gaussian likelihood \\
\midrule

\multicolumn{3}{l}{\textbf{Classification} (Two Moons)} \\
noise & Observation noise & 0.1 \\
epochs & Number of training epochs & 30 \\
batch\_size & Batch size & 25 \\
num\_samples & Number of training samples & 100 \\
num\_val\_samples & Number of validation samples & 100 \\
lr & Learning rate & 0.1 \\
optimiser & Optimiser & Adam \\
hidden\_dim & Hidden layer size & 10 \\
Divergence measure & The divergence measure used in this task & KL divergence, Categorical logits \\
\midrule

\multicolumn{3}{l}{\textbf{Generative Modelling} (Two Moons)} \\
batch\_size & Number of samples per training batch & 2,500 \\
epochs & Number of training epochs & 250 \\
hidden\_dim & Number of hidden units in the model & 64 \\
lr & Learning rate for optimiser & 0.01 \\
optimiser & Optimiser used for training & Adam \\
noise & Noise added to dataset & 0.05 \\
num\_integration\_steps & Number of integration steps in the model & 100 \\
num\_samples & Total number of training samples & 10,000 \\
num\_val\_samples & Number of validation samples & 1,000 \\
Divergence measure & The divergence measure used in this task & MMD with RBF kernel \\
\midrule
\bottomrule
\end{tabularx}
\label{tab:hyperparameters}
\end{table}

\end{document}